\documentclass[11pt]{article}

\usepackage[preprint]{acl}
\usepackage{amsmath,amsfonts,bm}

\def\eqref#1{equation~\ref{#1}}
\def\1{\bm{1}}

\def\rve{{\mathbf{e}}}

\def\rvu{{\mathbf{i}}}

\def\rvu{{\mathbf{u}}}
\def\rvv{{\mathbf{v}}}

\def\rvx{{\mathbf{x}}}
\def\rvy{{\mathbf{y}}}
\def\rvz{{\mathbf{z}}}

\def\vv{{\bm{v}}}

\def\vx{{\bm{x}}}

\DeclareMathAlphabet{\mathsfit}{\encodingdefault}{\sfdefault}{m}{sl}
\SetMathAlphabet{\mathsfit}{bold}{\encodingdefault}{\sfdefault}{bx}{n}

\usepackage{tabularx}
\usepackage{booktabs}
\usepackage{times}
\usepackage[most]{tcolorbox} % 启用 tcolorbox 全部常用功能
\usepackage{xcolor}          % 颜色支持
\usepackage{latexsym}
\usepackage{multirow}
\usepackage{amsmath}
\usepackage[T1]{fontenc}
\usepackage[utf8]{inputenc}

\usepackage{microtype}
\usepackage{amsfonts}
\usepackage{inconsolata}
\usepackage{subcaption}    % subfigure 环境
\usepackage{graphicx}
\title{From Static Personal Values to Contextualized Personalization: Bayesian Personalized Value Alignment for LLMs}

\author{
  \textbf{Hanze Guo\textsuperscript{1}}\thanks{Equal contribution.},
  \textbf{Aixuan Song\textsuperscript{1}}\footnotemark[1],
  \textbf{Jing Yao\textsuperscript{2}}\footnotemark[1],
  \textbf{Xiangxu Zhang\textsuperscript{1}},
\\
  \textbf{Xiaoyuan Yi\textsuperscript{2}},
  \textbf{Xing Xie\textsuperscript{2}},
  \textbf{Xiao Zhou\textsuperscript{1,3,4}}\thanks{Corresponding author.}
\\
\\
  \textsuperscript{1}Gaoling School of Artificial Intelligence, Renmin University of China
\\
  \textsuperscript{2}Microsoft Research Asia
\\
  \textsuperscript{3}Beijing Key Laboratory of Research on Large Models and Intelligent Governance
\\
  \textsuperscript{4}Engineering Research Center of Next-Generation Intelligent \\ Search and Recommendation, MOE
\\
  \texttt{\{ghz,songaix,xiaozhou\}@ruc.edu.cn, jingyao@microsoft.com}
}
\begin{document}
\maketitle
\begin{abstract}
% background -- problem/challenge -- core idea -- result
%Personalized value alignment for Large Language Models (LLMs) has become a practical demand to satisfy diverse user preferences.However, existing methods typically adapt model outputs across various prompts to a static value profile, ignoring how diverse contexts dynamically reshape the salience of different values.According to psychological theories like Lewin's Field Theory, human behavior is jointly shaped by personal values and situational constraints, implying that personal values act as merely priors while context-dependent posterior preferences drive ultimate behaviors.Thus, we propose \textbf{BaCVA}, an inference-time \textbf{Ba}yesian \textbf{C}ontext-aware personalized \textbf{V}alue \textbf{A}lignment method that dynamically approximates posterior personalized preference by integrating static personal values with scenario value salience. Specifically, it first estimates the value salience for a given context based on generally normative responses, then builds a \emph{dual-view personalization module} to infer posterior preferences from complementary personal-value and scenario-driven perspectives.BaCVA enables more accurate and adaptive personalized value alignment, as well as improved data efficiency by using prior information. Extensive experiments on two benchmarks showcase the superiority of BaCVA.

Personalized value alignment has become increasingly important as large language models (LLMs) are expected to accommodate diverse user preferences. However, existing methods typically align model outputs with a static value profile across prompts, overlooking that the salience of value dimensions varies substantially across contexts. Inspired by Lewin's Field Theory, which views human behavior as jointly shaped by personal dispositions and situational constraints, we model personal values as priors and context-dependent preferences as posteriors. We propose \textbf{BaCVA}, an inference-time \textbf{Ba}yesian \textbf{C}ontext-aware personalized \textbf{V}alue \textbf{A}lignment method that approximates posterior personalized preferences by integrating static personal values with scenario-specific value salience. BaCVA first estimates contextual value salience from generally normative responses, and then employs a \emph{dual-view personalization module} to infer posterior preferences from complementary personal-value and scenario-driven perspectives. This Bayesian formulation enables more accurate and adaptive personalized value alignment while improving data efficiency via prior values. Extensive experiments on benchmarks demonstrate its superiority over strong baselines.

\end{abstract}

\section{Introduction}\label{sec:intro}
The widespread deployment of Large Language Models (LLMs)~\citep{adler2024gpt, touvron2023llama, guo2025deepseek, yang2025qwen3,zhang2026r2medbenchmarkreasoningdrivenmedical,
li-etal-2025-automir,
zhu2026mohobench,
zhang-etal-2026-inflated,
DBLP:conf/nips/YongZZLZW25,
DBLP:conf/acl/YongLY0025,
DBLP:conf/www/YongSWZ26} has made aligning LLMs with human values a central research challenge~\citep{wang2023aligning_survey, shen2023align_survey, yao2023align_survey, wang2024essence,guo2026counterfactual,
zhang2026humanvaluesmatterinvestigating}. While prominent methods focus on promoting universal but monolithic values, such as \emph{HHH} (helpful, harmless, honest)~\citep{bai2022align_hhh, ouyang2022training, hendrycks2020ethics}, they struggle to accommodate the substantial diversity of individual user values. Consequently, \emph{personalized value alignment} emerges as a more practical objective~\citep{zhou2025tricolore,
guo2025sorex,
guo2026not,
zhang-etal-2026-hypemed}.

Existing work on personalized value alignment falls into two categories. \emph{Training-time methods} fine-tune personalized reward models and adapters using user-specific interaction data~\citep{poddar2024personalizing,zhao2025teaching}. Nevertheless, these methods typically rely on extensive personal value data and might degrade in data-scarce or cold-start scenarios~\citep{puadurean2025inference,poddar2024personalizing}. \emph{Inference-time methods} offer greater flexibility, which steer model outputs by personal value prompts~\citep{puadurean2025inference,ryan2025synthesizeme,chen2025popi}, activating value-related model parameters~\citep{zhu2024personality,zhang2025personalize} or modifying token distributions during decoding~\citep{chen2024pad,shi2024decoding,zhang2025persona}.
However, \emph{most inference-time methods face a common limitation}: they conceptualize personalized value alignment as adapting model outputs to a \emph{static personal value profile} to capture users' overall value orientations. \emph{Although these methods condition response generation on the input question, they typically use the same user value profile for every question, without explicitly estimating which value dimensions become more or less important in the current context.}

% generic and static value statements provided by individual users. Though such descriptions may capture a user's overall value orientation, \textit{they overlook how diverse and nuanced scenarios could impact the salience and prioritization of values}, leading to sub-optimal alignment with personalized values.

% To address personalized value alignment, existing research has proposed several methods that attempt to align LLMs at individual, group, and cultural levels through techniques such as model fine-tuning or context learning \citep{jang2023personalized, lee2024aligning, masoud2023cultural, li2024culturellm}. 

% However, these methods often overlook the importance of scenario alignment. 

\begin{figure*}[t]
\centering

\begin{subfigure}[t]{0.70\linewidth}
    \centering
    \includegraphics[width=\linewidth]{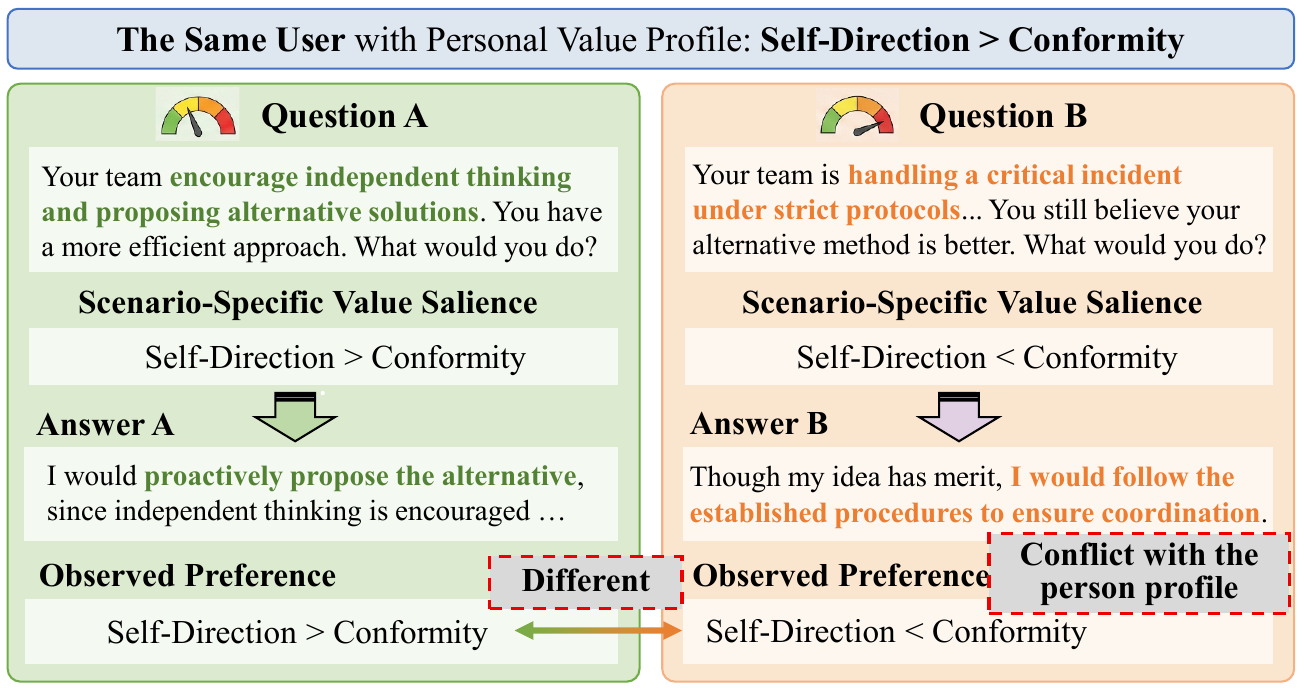}
    \caption{}
\end{subfigure}
\hfill
\begin{subfigure}[t]{0.29\linewidth}
    \centering
    \includegraphics[width=\linewidth]{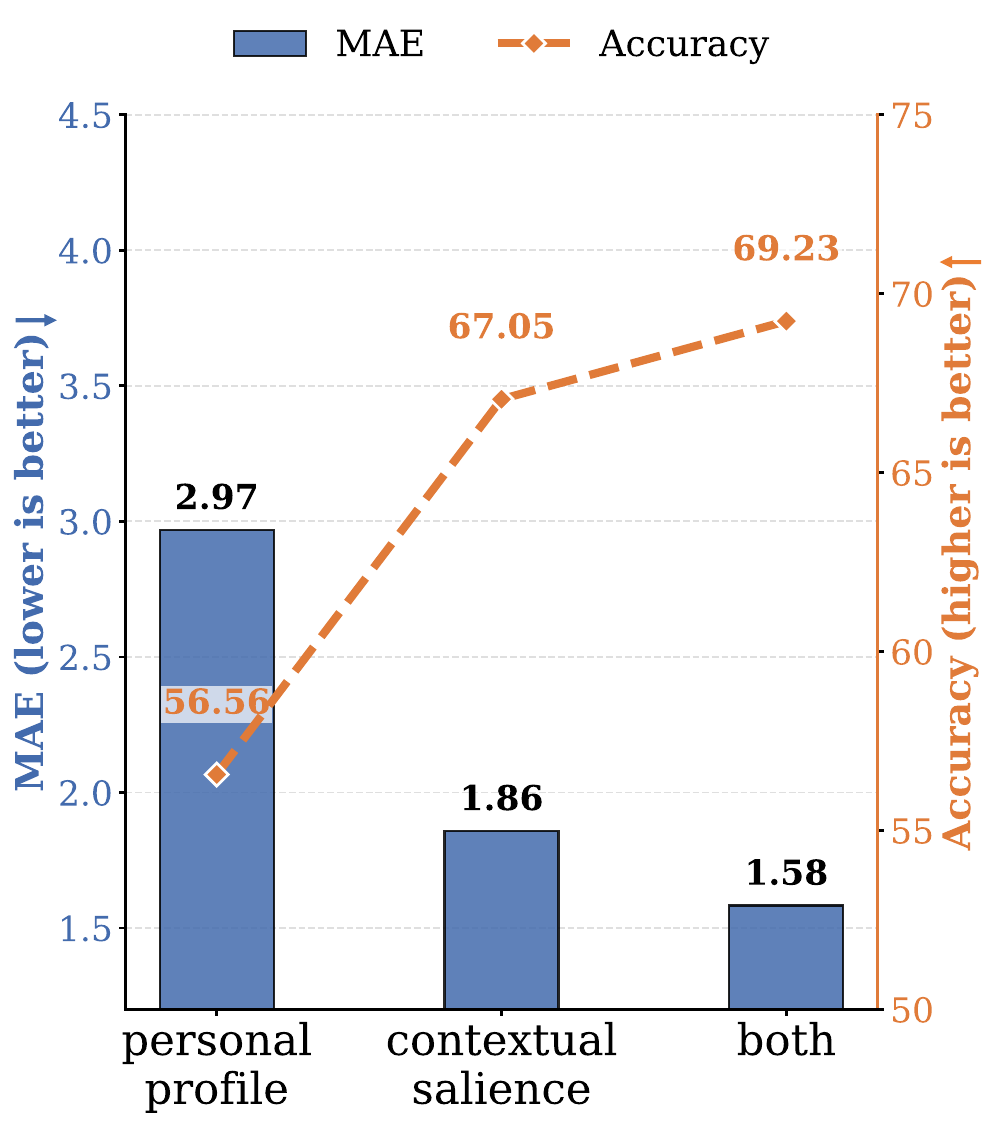}
    \caption{}
\end{subfigure}

\caption{
(a) An example showing that a user a stable personal value profile may exhibit different value preferences across contexts, which are jointly shaped by personal values and situational constraints. (b) Empirical results show that only personal values or scenario value salience achieve weaker personalization than their combination.
% Context-Aware Value Alignment Across Scenarios. A user may exhibit different value preferences under different contexts, and combining user-level priors with contextual signals (b) leads to improved alignment.
}
\label{fig:prelim_analysis}
\end{figure*}

Drawing from well-established psychological theories~\citep{mischel2013personality, epstein1985person}, stable internal dispositions alone are insufficient to explain human behaviors across varying contexts. Notably, Lewin’s Field Theory~\citep{heidbreder1937lewin} posits that observed behavior is jointly shaped by personal values and situational constraints.
As illustrated in Fig.\ref{fig:prelim_analysis} (a), an individual who inherently prioritizes \textit{`Self-Direction'} over \textit{`Conformity'} may nonetheless suppress self-direction actions in scenarios governed by other values. This suggests that each scenario may selectively activate or suppress different value dimensions according to broadly shared social norms~\citep{cialdini1990focus,burnes2004kurt}, which can be formalized as a context-aware \emph{value salience distribution}. Together, inherent personal values and context-aware value salience ultimately shape behavior. This is also supported by empirical results in Fig.~\ref{fig:prelim_analysis} (b) that combining both leads to more accurate preference prediction. From a Bayesian lens, inherent personal values and contextual value salience serve as complementary \emph{priors}, while the \emph{posterior values} inferred from observed behaviors reflect final context-aware personal preferences. 

% As illustrated in Fig.\ref{fig:prelim_analysis} (a), an individual who stably prioritizes \textit{`Self-Direction'} over \textit{`Conformity'} may nonetheless suppress self-direction actions in scenarios where competing values are much more salient. This example implies that each scenario induces a \emph{value salience distribution} shaped by universal social norms, where some values are encouraged while others are less acceptable~\citep{cialdini1990focus,burnes2004kurt}. The values that ultimately guide decision-making emerge from the interaction between the stable personal values and context-specific value salience (empirically supported by Fig.~\ref{fig:prelim_analysis} (b)). From a Bayesian probabilistic lens, the static personal value profile and context-specific value salience serve as complementary \emph{priors} over latent values for decision-making, while the \emph{posterior values} inferred from observed behaviors reflect context-aware personal preferences. 
% Therefore, either the personal value profile or context-specific value salience alone can only serve as prior conditions of latent value representations for decision-making, and the posterior values inferred from the observed user behaviors are exactly what we need to learn from.

%% revised by jy for EMNLP
Inspired by this formulation, we propose \textbf{BaCVA}, an inference-time \textbf{Ba}yesian \textbf{C}ontext-aware personalized \textbf{V}alue \textbf{A}lignment method. It achieves personalization by dynamically approximating the posterior personalized values based on both static personal values and context-specific value salience.
Specifically, BaCVA consists of two components: (i) a \textit{scenario value estimator} that predicts the context-aware value salience; (ii) a \textit{dual-view personalization module} that approximates posterior personal values from complementary personal value-driven and scenario-driven perspectives. The whole framework is optimized toward a variational inference objective~\cite{tzikas2008variational,blei2017variational}. %anchored in either the proir information of context-specific values salience or static personal values. The whole framework is trained with a variational inference objective that maximizes an Evidence Lower Bound (ELBO).
By framing personalization as context-aware posterior inference, BaCVA offers two advantages: (i) \emph{fine-grained adaptability}, capturing nuanced value shifts from priors across diverse scenarios; and (ii) \emph{data efficiency}, enabling personalization with fewer training samples and seamless generalization to new users. Extensive experiments on multiple benchmarks showcase the superiority of our framework.

The main contributions are summarized below. \textbf{(1)} We theoretically and empirically identify a key limitation in current personalized alignment methods: they typically reuse a fixed personal value profile across prompts without explicitly modeling question-specific changes in value salience. \textbf{(2)} We propose BaCVA, an inference-time Bayesian personalized alignment framework. It achieves dynamic personalization by inferring context-aware posterior preferences conditioned on prior personal values and contextual value salience. \textbf{(3)} Extensive experiments show that BaCVA achieves more accurate and adaptive personalized value alignment, as well as improved data efficiency and generalization.

\section{Related Work}

% \subsection{General Value Alignment of LLMs}
\paragraph{General Value Alignment of LLMs}
General (non-personalized) value alignment aims to align LLM outputs with population-level human preferences like HHH (helpfulness, harmlessness and honesty)~\citep{wang2023aligning_survey, shen2023align_survey, yao2023align_survey}. Prominent methods include supervised fine-tuning (SFT)~\citep{zhang2023instruction,han2025towards}, reinforcement learning from human feedback (RLHF)~\citep{ouyang2022training, bai2022align_hhh}, direct preference optimization (DPO)~\citep{rafailov2023direct,ethayarajh2024kto,meng2024simpo} and variants. Besides, inference-time methods like retrieval-augmented generation (RAG)\citep{lewis2020retrieval,zhao2024retrieval,gao2023retrieval} and principle-based prompting\citep{bai2022constitutional} steer model output without parameter updates.

These methods inherently struggle to accommodate the diversity of individual user values, motivating the need for personalized value alignment. % Besides, the population-level norms captured in these methods imply the scenario-specific value salience, acting as a prior in our framework.

% \subsection{Personalized Value Alignment of LLMs}
\paragraph{Personalized Value Alignment of LLMs}
Existing methods fall into two categories. \emph{Training-time personalization} incorporates user preferences via model optimization~\citep{chen2025pal,choi2025copl}. VPL~\citep{poddar2024personalizing} and RLPA~\citep{zhao2025teaching} learn user-specific reward models or adapters from individual interaction data. P-RLHF~\citep{li2024personalized} learns user embeddings for personalization. 
Despite these methods implicitly fit to observed posterior preferences, they rely on extensive per-user data and degrade in cold-start or sparse-data settings~\citep{puadurean2025inference}.
% Although effective with sufficient supervision, these approaches rely heavily on per-user value annotations and degrade in cold-start or sparse-feedback settings~\citep{puadurean2025inference}, and the resulting personalization is largely embedded into model parameters, limiting flexibility.

In contrast, \emph{inference-time personalization} steers model behaviors by manipulating prompts~\cite{chen2025popi,jincontext} or decoding~\citep{thonet2025fast,zhu2024personality}, without modifying parameters. ValuePrompt~\citep{kang2024causal_align} and COUPLE~\citep{guo2026counterfactual} condition generation on explicit user value descriptions. Other approaches summarize or retrieve value-aware content from user histories, like ValuesRAG~\citep{ryan2025synthesizeme,seo2025valuesrag}. 
Decoding-based methods~\citep{zhang2025persona} like MOD~\citep{shi2024decoding} and PAD~\citep{chen2024pad} reshape token distributions to adapt general outputs to fixed individual preferences.
These inference-time approaches mainly reuse the same personal value profile across situations, without explicitly modeling question-specific changes in the salience of individual value dimensions. We address this limitation in this paper.

% Despite their flexibility and interpretability, 
% In this paper, we address this challenge by grounding personalization on the final observed personal preferences under each scenario.
% approximating the final observed personal preferences under each scenario with the static personal value and scenario information as the priors.

% \subsection{Bayesian Modeling and Applications}
\paragraph{Bayesian Modeling and Applications}
Bayesian modeling~\citep{zhang1996exploitingcausalindependencebayesian, bishop2013variationalrelevancevectormachines, Grossman2004LearningBN, Lauritzen1990LocalCW, heckerman2015learningbayesiannetworkscombination} offers a principled lens for updating prior beliefs with observed evidence and reasoning with uncertainty~\citep{Gao_2024}. Recent work~\citep{gupta2025coinflipsmakellms,longjohn2025bayesian,wang2024aligninglanguagemodelshuman} has explored LLMs alignment from a Bayesian perspective. BIRD~\citep{feng2025birdtrustworthybayesianinference} frames decision alignment as a Bayesian decision process. Dellma~\citep{liu2024dellma} applies decision theory to address uncertainty.
% addresses decision-making under uncertainty grounded in decision theory. % by introducing multi-step reasoning 

% BIRD~\citep{feng2025birdtrustworthybayesianinference} frames decision alignment as a Bayesian decision process with uncertainty modeling, preference representation, and belief updating. Dellama~\citep{liu2024dellma} addresses decision-making under uncertainty by introducing multi-step reasoning grounded in  decision theory.

This paper formulates personalized value alignment as Bayesian posterior inference, optimized through Variational Inference~\citep{tzikas2008variational,blei2017variational,ganguly2021introduction}.

% This paper applies Bayesian posterior inference to formulate personalized value alignment, optimized through Variational Inference~\citep{tzikas2008variational,blei2017variational,ganguly2021introduction}.

% In contrast to Bayesian approaches that primarily target population-level heterogeneity, our work uses Bayesian inference as a \emph{mechanism for personalized decision integration}: we treat a user’s stable value profile as a prior and perform context-conditioned posterior updates to obtain scenario-aware personalized value orientations that guide aligned responses.
% \input{section_revised/preliminary}
\section{Methodology}
\label{sec:method}

\begin{figure*}[!ht]
    \centering
    \includegraphics[width=\linewidth]{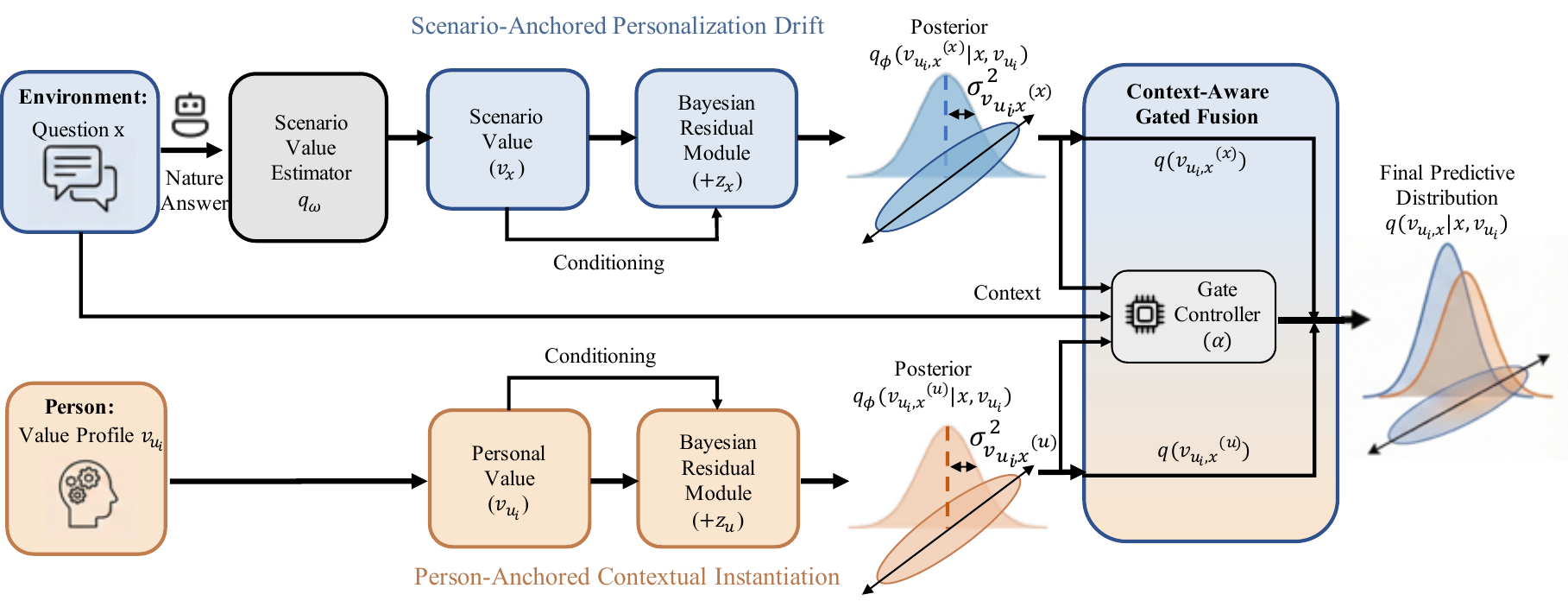}
    \vspace{-3mm}
    \caption{
    Overview of the proposed Bayesian question-specific value modeling framework.
    A question $x$ provides a decision context, while a user-level value profile $v_{u_i}$
    modulates how semantics are personalized. We learn a Bayesian \emph{residual} (semantic delta) from two complementary views, and fuse the two views via a context-based gated fusion mechanism to predict a personalized, context-aware answer.
    }
    \vspace{-3mm}
    \label{fig:model_overview}
\end{figure*}

% We predict a personalized answer representation
    % by learning a Bayesian \emph{residual} (semantic delta) around a universal anchor.

\subsection{Task Definition}
\label{subsec:task_definition}
% This paper concerns personalized value alignment of LLMs. 
Let $\mathcal{U} = \{u_i\}_{i=1}^{N}$ denote a population of diverse users, each user $u_i$ is associated with a self-written summary of personal values $c_{u_i}$ and a small set of preference annotations $\mathcal{D}_{u_i}=\{(x,y_{c},y_{r})\}$, where $x$ denotes a prompt or question, $y_{c}$ and $y_{r}$ mean the preferred and dispreferred responses respectively. To make the open-ended value summary $c_{u_i}$ more actionable and unified, we follow~\cite{yao2023value_fulcra,guo2026counterfactual} to convert it into a structured, multi-dimensional value profile $\bm v_{u_i} = [(v_1,s_1),\ldots,(v_d,s_d)]$. $(v_1, \ldots, v_d)$ are finite pre-defined value dimensions (e.g., care, authority) from a well-established value framework (e.g., Schwartz's Basic Values~\citep{schwartz2012overview}). $s_j \in \{1,2,3,4,5\}$ indicates the priority assigned to $v_j$ on a 5-point Likert scale from \emph{`not important at all (1)'} to \emph{`very important (5)'}.

Since training a dedicated model for each user is less practical in low-data (small $\mathcal{D}_{u_i}$) and cold-start cases, we focus on inference-time personalized alignment in this paper. With a single shared model $p_{\psi}(\rvy | \rvx, \rvv_{u_i,\bm x})$ that can generate the response $\rvy$ aligned with the given personalized values $\rvv_{u_i,\bm x}$ toward the question $\rvx$, our goal is to infer the concrete personalized values $\rvv_{u_i, x} \sim p_{\theta}(\rvv|\rvv_{u_i}, \rvx)$ the user would exhibit to guide response generation under the specific context $\rvx$.

For evaluation on test samples $\{(u_i, x, y_c, y_r)\}$, we extract the values reflected in two responses as $\vv_{y_c}, \vv_{y_r}$, then compare inferred personalized values $\vv_{u_i, x}$ with them using two metrics: (i) the distance between $\vv_{u_i, x}$ and $\vv_{y_c}$ like MAE and (ii) the accuracy whether $\vv_{u_i, x}$ is closer to $\vv_{y_c}$ than to $\bm v_{y_r}$.
Furthermore, we evaluate the effectiveness of BaCVA in guiding downstream personalized response generation. We compare the responses generated from the inferred value $\vv_{u_i, x}$ against $y_c$ and $y_r$.

% We consider three key representations.
% $V_P$ denotes a user's intrinsic value profile (persona).
% $v_{uni}$ is the universal answer anchor produced by a frozen LLM,
% representing a generic, non-personalized baseline.
% $v_a$ is the ground-truth personalized answer embedding.
% Our analysis examines how $v_a$ deviates from $V_P$ and $v_{uni}$,
% and how these signals complement each other to motivate our model
% (see Figure~\ref{fig:prelim_analysis}).

\subsection{The BaCVA Framework}
\label{subsec:framework}
As discussed in Sec.\ref{sec:intro}, existing inference-time personalization methods employ the overall user profile $\vv_{u_i}$ or that summarized from the preference data $\vv_{u_i, \mathcal{D}_i}$ to instantiate $\rvv_{u_i, x}$ and keep it fixed across prompts $\rvx$, thereby limiting their ability to explicitly model question-specific shifts in the salience of individual value dimensions, as illustrated in Fig.~\ref{fig:prelim_analysis} (a).
% failing to address the challenge that value salience is context-dependent and affects observed behaviors, as illustrated in Fig.~\ref{fig:prelim_analysis} (a).

% Inspired by Lewin's field theory~\citep{heidbreder1937lewin} that observed behavior emerges from the interaction between the person (personal value profile $\vv_{u_i}$) and the situation (question $\rvx$), this paper proposes BaCVA, a Bayesian framework that formalizes personalized value alignment as a posterior inference task by explicitly integrating both the personal value profile and specific context as prior information. Given the observed preference data $\mathcal{D}_{u_i}=\{(x,y_{c},y_{r})\}$, the posterior distribution of the target $\rvv_{u_i,i}$ we want to learn is formalized as: 
% \begin{align}
%     & p_{\theta}(\rvv_{u_i,x}|\rvv_{u_i},\rvx,y_c \succ y_r) \\ \nonumber
%     & \propto p_{\theta}(y_c \succ y_r|\rvv_{u_i,x})\times p_{\theta}(\rvv_{u_i,x}|\rvv_{u_i}, \rvx).
% \end{align}

%% revised by jy for EMNLP
Inspired by Lewin's field theory~\citep{heidbreder1937lewin} that observed behaviors emerge from the interaction between the personal value profile $\rvv_{u_i}$ and the situation $\rvx$, and Cognitive Appraisal Theory~\citep{lazarus1984stress} that situations modulate the expression of internal values to form final actions, we formalize the contextual factor as a scenario-specific value salience distribution. By treating both the personal value profile and specific context value salience as priors (prior to behavioral execution), we propose BaCVA, a Bayesian framework that dynamically estimates the contextual personalized values $\rvv_{u_i,x}$ via posterior inference. Given the data $\mathcal{D}_{u_i}=\{(x,y_{c},y_{r})\}$, we model observed user preferences $y_c \succ y_r$ as the likelihood and the posterior distribution of the target $\rvv_{u_i,x}$ is formalized as a calibration from the prior value distribution $p_{\theta}(\rvv_{u_i,x}|\rvv_{u_i}, \rvx)$ by the likelihood:
\begin{align}\label{eq:posterior}
    & p_{\theta}(\rvv_{u_i,x}|\rvv_{u_i},\rvx,y_c \succ y_r) \\ \nonumber
    & \propto p_{\theta}(y_c \succ y_r|\rvv_{u_i,x})\times p_{\theta}(\rvv_{u_i,x}|\rvv_{u_i}, \rvx).
\end{align}

% Given the observed preference data $\mathcal{D}_{u_i}=\{(x,y_{c},y_{r})\}$,
% \textbf{we treat the preference event $o:=(y_c>y_r)$ as indicating the chosen response $y_c$,
% from which we extract a value-level observation $\rvv_a := f_{\mathrm{val}}(y_c)$.
% The posterior distribution of the target $\rvv_{u_i,x}$ is then formalized as:}
% \begin{align}
%     p_{\theta,\phi}(\rvv_{u_i,x}\mid \rvv_{u_i},\rvx,\rvv_a)
%     \propto
%     p_{\theta}(\rvv_a\mid \rvv_{u_i,x},\rvx)\times p_{\phi}(\rvv_{u_i,x}\mid \rvv_{u_i}, \rvx).
% \end{align}

% This target corresponds to a calibration from the prior value distribution $p_{\theta}(\rvv_{u_i,x}|\rvv_{u_i}, \rvx)$ through observed behaviors. 
This formalization yields two advantages: (1) more adaptive personalization by enabling fine-grained calibration from informative priors, and (2) training efficiency by introducing high-quality priors. Moreover, BaCVA does not rely on fine-tuning user-specific adapters but personal value profiles, thus easily extended to cold-start users.

Since direct computation of the posterior value distribution is intractable, we refer to Variational Inference (VI)~\citep{blei2017variational} and introduce a variational posterior $q_{\phi}(\rvv_{u_i,x}|\rvv_{u_i},\rvx)$ implemented by LLMs. As shown in Fig.\ref{fig:model_overview}, the architecture consists of two modules: (i) a \textit{scenario value estimator} to predict contextual value salience; and (ii) a \textit{dual-view personalization module} that approximates the posterior distribution from both personal value-driven and scenario-driven perspectives. Then, we optimize the model by minimizing the divergence between the variational posterior and the true posterior, which is equivalent to maximizing the evidence lower bound (ELBO) below, with full derivation in Appendix~\ref{app:generic_elbo}.
% \begin{align}\label{eq:elbo_loss}
%     \mathcal{L} = & -\mathbb{E}_{q_{\phi}(\rvv_{u_i,x})}[\log p_{\theta}(\rvv_{u_i,x}|v_{u_i}, x, y_c \succ y_r)] \\\nonumber
%     & + \text{KL}(q_{\phi}(\rvv_{u_i,x})||p_{\theta}(\rvv_{u_i,x}|\rvv_{u_i}, \rvx))
% \end{align}
\begin{align}\label{eq:elbo_loss}
\mathcal{L}
= &-\mathbb{E}_{q_{\phi}(\rvv_{u_i,x})}\!\left[\log p_{\theta}(y_c \succ y_r \mid \rvv_{u_i,x}, x)\right] \\ \nonumber
&+ \mathrm{KL}\!\left(q_{\phi}(\rvv_{u_i,x}\mid \rvv_{u_i}, \rvx) \,\|\, p_{\theta}(\rvv_{u_i,x}\mid \rvv_{u_i}, \rvx)\right).
\end{align}

% \begin{align}\label{eq:elbo_loss}
% \mathcal{L}
% = &-\mathbb{E}_{q_{\phi}(\rvv_{u_i,x})}\!\left[\log p_{\theta}(v_a \mid \rvv_{u_i,x}, x)\right] \\
% &+ \mathrm{KL}\!\left(q_{\phi}(\rvv_{u_i,x}) \,\|\, p_{\theta}(\rvv_{u_i,x}\mid \rvv_{u_i}, \rvx)\right).
% \end{align}

Each module is detailed in the next subsections.

\subsection{Estimation of Scenario Value Salience}\label{subsec:scenario_value_salience}
As illustrated in Fig.~\ref{fig:prelim_analysis}, different value dimensions are not equally salient across scenarios, instead, a specific question $x$ usually activates only a subset of dimensions. Typically, the salience distribution over these dimensions are shaped by universal social norms, encouraging some values while rendering others less acceptable.
% Furthermore, the question induces a default salience distribution on these dimensions shaped by universal social norms, which makes some values more encouraged while rendering others less acceptable.

Inspired by this, we estimate the scenario value salience distribution $\rvv_x$ based on population-level aligned response. Concretely, for a question $x$, we leverage LLMs that have gone through great human alignment as a proxy to obtain the universal response, denoted as $\rvy_{\text{human}}$. Then, we employ a prompt-based value estimator $q_{\omega}$ to infer $\rvv_x$:
\begin{align}
    \rvv_x \sim q_{\omega}(\cdot|\rvx, \rvy_{\text{human}}).
\end{align}

In our primary implementation, we leverage GPT-5-nano to generate the universal response and estimate the value salience through in-context learning. We also replace it with other LLMs to validate the robustness in Appendix~\ref{app:addition_exp}. Detailed prompts are provided in Appendix~\ref{Methods_prompt}.

% For each question $Q$, we obtain a generic, population-level response value
% \[
% v_{uni} = \mathrm{Enc}(\text{CanonicalAnswer}(Q))
% \in \mathbb{R}^d,
% \]
% \noindent which serves as a non-personalized semantic value baseline.

% \paragraph{Context--Value Interaction.}
% Not all value dimensions are equally relevant to every question. 
% We therefore construct a context-aware interaction representation $h_{cv}$ that 
% captures how the question semantically selects and weights user value dimensions:
% \begin{equation}
% h_{cv} = f_{\text{int}}(Q, V_P),
% \end{equation}
% where $f_{\text{int}}(\cdot)$ models how contextual semantics modulate value importance,
% yielding a compact representation of person--environment interaction.
% In practice, $f_{\text{int}}$ is instantiated using an LLM-based
% module that jointly processes the question and the value profile to
% produce a context-sensitive interaction representation; implementation
% details are provided in Appendix~\ref{app:method}.

\subsection{Dual-View Personalized Value Inference}\label{subsec:dual_view_personalization}
Lewin’s field theory states that behavior arises from the interaction between the person and the environment. Thus, the concrete values reflected in final behaviors can be explained from two complementary perspectives: (1) \emph{Scenario-anchored personalization drift}, where shared norms under the specific scenario largely determine the behavior while personal values introduce slight but meaningful deviation; (2) \emph{Person-anchored contextual instantiation}, which assumes that personalization is primarily rooted in the user's inherent value profile while the scenario modulates how salient each value is.
Motivated by this, we design a \textbf{dual-view personalization module} to approximate the posterior distribution over $\rvv_{u_i,\vx}$. Specifically, it includes a \emph{scenario-driven component} and a \emph{person-driven component} to infer the posterior respectively:
\begin{align}
    q_{\phi}(\rvv_{u_i,\vx}^{(\rvx)}|\rvv_{x},\rvv_{u_i}), \quad q_{\phi}(\rvv_{u_i,\vx}^{(\rvu)}|\rvv_{u_i}, \rvv_{x}).
\end{align}
Then, the final variational posterior is calculated as a gated composition of the two perspectives:
\begin{align}
    q_{\phi}(\rvv_{u_i,x}|\rvv_{u_i},\rvx) = \alpha q_{\phi}(\rvv_{u_i,\vx}^{(\rvx)}|\rvv_{x},\rvv_{u_i}) \\ \nonumber
    + (1-\alpha) q_{\phi}(\rvv_{u_i,\vx}^{(\rvu)}|\rvv_{u_i}, \rvv_{x}),
\end{align}
where $\alpha$ is produced through \emph{a context-aware gate}. % Each component is detailed in the following.

\paragraph{View 1: Scenario-Anchored Personalization Drift}
Conditioned on the scenario-based value salience prior $q_{\omega}(\rvv_x)$, this view models personalization as adding a personal residual to the scenario anchor in logit space:
\begin{align}
q_{\phi}(\rvv^{(x)}_{u_i,x}\mid \rvv_x,\rvv_{u_i})
&=
\mathrm{Softmax}(\mathbf{s}_x+z_u), \\
\mathbf{s}_x
&=
\mathrm{Logit}(q_{\omega}(\rvv_x)),
\end{align}
where $z_u$ is the personal residual logit vector. Intuitively, this view is effective when most individuals follow the scenario-level consensus with slight personalized divergence.

\paragraph{View 2: Person-Anchored Contextual Instantiation}
This view starts from the personal value prior $q_{\phi}(\rvv_u)$ and instantiates it under the current scenario by adding a context residual in logit space:
\begin{align}
q_{\phi}(\rvv^{(u)}_{u_i,x}\mid \rvv_{u_i},\rvv_x)
&=
\mathrm{Softmax}(\mathbf{s}_u+z_x), \\
\mathbf{s}_u
&=
\mathrm{Logit}(q_{\phi}(\rvv_u)),
\end{align}
where $z_x$ is the context residual logit vector. This perspective emphasizes personal value-driven behaviors where scenario-level expectations provide only weak constraints.

% This perspective emphasizes strong, value-driven deviations where the universal anchor provides little guidance, capturing cases where users' intrinsic values play a dominant role.
 
% We first obtain a latent persona representation $v_p$ and infer a Gaussian refinement around it:
% \begin{equation}
% z_p \sim p_{p}(z \mid h_{cv}), \qquad 
% \hat v_a^{(p)} = v_p + z_p.
% \end{equation}
% Unlike the universal-drift view, this perspective emphasizes strong,
% value-driven deviations where the universal anchor provides little guidance, capturing cases where users' intrinsic values play a dominant role.

In practice, we employ open-sourced LLMs to represent these distributions by prompting them with appropriate instructions and extracting the generation probability of each value. More details about the prompts are provided in Appendix~\ref{Methods_prompt}.

% In this paper, we implement this through an open-sourced LLMs by prompting it with the universal value anchor and ask it to adjust the values by capturing the residual. 

% \paragraph{Uncertainty-Aware Gated Fusion}
\paragraph{Context-Aware Gated Fusion}
The above dual-view posterior values are complementary and capture dominant information suited for different scenarios. Rather than equally combining them across all contexts, it would be more effective to dynamically consider their relative dominance. Consequently, we introduce a context-aware fusion mechanism to compute the gate score $\alpha \in [0,1]$ as:
\begin{align}
\scalebox{0.95}{$%
    \alpha = \mathcal{G}(\vx, q(\rvv_{u_i,\vx}^{(\rvx)}), q(\rvv_{u_i,\vx}^{(\rvu)}), \sigma^2_{\rvv_{u_i,\vx}^{(\rvx)}}, \sigma^2_{\rvv_{u_i,\vx}^{(\rvu)}}).
$}
\end{align}
$q(\rvv_{u_i,\vx}^{(\rvx)})$ is the abbreviation of $q(\rvv_{u_i,\vx}^{(\rvx)}|\rvv_{x},\rvv_{u_i})$, $\sigma^2_{\rvv_{u_i,\vx}^{(\rvx)}}$ and $\sigma^2_{\rvv_{u_i,\vx}^{(\rvu)}}$ are variances to measure the uncertainty of posterior inference in the two views. In practice, the gate function $\mathcal{G}$ corresponds to an LLM, which jointly reasons over the inputs and outputs a continuous context-aware fusion weight by normalizing the logits of tokens \texttt{0} and \texttt{1}, rather than making a hard binary decision. Specific implementation details are given in Appendix~\ref{Methods_prompt}. % while the overall formulas of our framework are provided in Appendix~\ref{app:formula}.

\paragraph{Variational Inference Optimization} 
We use the same value extractor in Subsec.~\ref{subsec:scenario_value_salience} to analyze the value reflected by the preferred response $v_{y_c}$, used as the supervision during training. Besides, we approximate the expectation term in Eq.(\ref{eq:elbo_loss}) with a single-pass amortized estimate, getting the predicted posterior value as $\hat v_{u_i,x}$. Then, the Variational Inference loss in our dual-view personalization model is:
% \begin{align}\label{eq:dual_loss}
%     \mathcal{L} = \left\|\hat v_{u_i,x} - v_{y_c}\right\|_2^2+
% \lambda_1 \sum_{k\in\{\mathrm{u},\mathrm{x}\}} \mathrm{KL}(q_{\phi}^{(k)} \,\|\, p_{\theta}^{(k)}).
% \end{align}
\begin{equation}\label{eq:dual_loss}
\scalebox{0.82}{$%
    \mathcal{L} = \left\|\hat v_{u_i,x} - v_{y_c}\right\|_2^2+
\lambda_1 \sum_{k\in\{\mathrm{u},\mathrm{x}\}} \mathrm{KL}(q_{\phi}^{(k)} \,\|\, p_{\theta}^{(k)}).
$}
\end{equation}

% We use $v_a = f_{\text{eval}}(y_c)$ (details in Appendix~\ref{Preprocessing_prompt} and human evaluated in~\ref{app:human_eval}) as the evaluation metric during training to determine whether $y_c \succ y_r$. Therefore, the optimization objective changes from
% \begin{equation}
% -\mathbb{E}_{q_{\phi}(v_{u_i}, x)}
% \left[
% \log p_{\theta}(y_c \succ y_r \mid v_{u_i}, x)
% \right]
% \end{equation}
% to
% \begin{equation}
% -\mathbb{E}_{q_{\phi}(v_{u_i}, x)}
% \left[
% \log p_{\theta}(v_a \mid v_{u_i}, x)
% \right].
% \end{equation}

% In our proposed dual-view personalization model, the Variational Inference loss in Eq(\ref{eq:elbo_loss}) is instantiated as:
% \begin{align}\label{eq:dual_loss}
%     \mathcal{L} = \underbrace{\left\|\hat v_a - v_a\right\|_2^2}_{\mathcal{L}_{\text{rec}}}+
% \lambda_1\,
% \underbrace{
% \sum_{k\in\{\mathrm{u},\mathrm{x}\}}
% \mathrm{KL}(q_k \,\|\, p_k)
% }_{\mathcal{L}_{\mathrm{KL}}} .,
% \end{align}
% which is constrained by the dual-view prior conditions.

% In practice, we combine the ELBO with a contrastive loss to enhance discriminability. The total objective is:
% \begin{equation}
% \mathcal{L}
% =
% \underbrace{\left\|\hat v_a - v_a\right\|_2^2}_{\mathcal{L}_{\text{rec}}}
% +
% \lambda_1 \underbrace{\sum_{k \in \{uni, p\}} \mathrm{KL}(q_k \| p_k)}_{\mathcal{L}_{\text{KL}}}
% +
% \lambda_2 \mathcal{L}_{\text{NCE}},
% \label{eq:loss_total}
% \end{equation}
% where $\mathcal{L}_{\text{NCE}}$ is the InfoNCE loss aligning $\hat v_a$ with $v_a$ against in-batch negatives using cosine similarity (implementation details are provided in Appendix~\ref{app:method}).
\section{Experiments}

\begin{table*}[ht]
\centering
\small
\renewcommand{\arraystretch}{1.3}
\setlength{\tabcolsep}{6pt}
\begin{tabular}{l l c c c c c}
\hline
 &  & \multicolumn{3}{c}{\textbf{PRISM}} & \multicolumn{2}{c}{\textbf{GOOD}} \\
\cline{3-5} \cline{6-7}
\textbf{Category} & \textbf{Method} 
& \textbf{MAE} $\downarrow$ 
& \textbf{Correlation} $\uparrow$ 
& \textbf{Accuracy} $\uparrow$
& \textbf{MAE} $\downarrow$
& \textbf{Correlation} $\uparrow$ \\
\hline

\multirow{1}{*}{Non-personalized Alignment} 
% & Q-Only 
% & 3.519 & 0.109 & 48.92
% & 3.422 & -0.099 \\

& DirectAnswer
& 1.858 & 0.485 & 67.05
& 3.302 & 0.517 \\
\hline

\multirow{8}{*}{\shortstack{Inference-time\\Personalized Alignment}}
& PersonValue 
& 2.969 & 0.411 & 56.56
& 3.698 & 0.189 \\

& Value Prompt 
& 1.790 & 0.568 & 72.95
& 3.062 & 0.579 \\

& MetaAligner 
& 1.955 & 0.509 & 69.00
& 3.231 & 0.480 \\

& COUPLE 
& 1.819 & 0.588 & 65.98
& 3.089 & 0.347 \\

& PAD 
& 1.733 & 0.607 & 66.67
& 3.236 & 0.360 \\

& MOD 
& 1.620 & 0.595 & 66.37
& 3.120 & 0.410 \\

& ValuesRAG 
& \underline{0.952} & \underline{0.638} & \underline{71.68}
& \underline{2.261} & \underline{0.589} \\
\cline{2-7}

& \textbf{BaCVA (Ours)}
& \textbf{0.728}$^{*}$ 
& \textbf{0.700}$^{*}$ 
& \textbf{81.72}$^{*}$
& \textbf{1.118}$^{*}$ 
& \textbf{0.769}$^{*}$ \\
\hline
\end{tabular}
\caption{Main results on \textsc{PRISM} and \textsc{GOOD}. The best results are bold and second-best results are underlined. $^{*}$ indicates statistically significant improvement over all baselines ($p < 0.05$). Accuracy is not applicable to GOOD that has only a ground truth answer but not preferred and dispreferred response pairs.}
% Lower MAE is better; higher Correlation and Accuracy are better (Accuracy is not applicable to GOOD). 
% $^{*}$ indicates statistically significant improvement over all baselines ($p < 0.05$).}
\label{tab:main_prism_good}
\end{table*}

\subsection{Experiment Setting}

% \paragraph{Datasets.}
% We conduct evaluation on two benchmarks, \textbf{PRISM}~\citep{kirk2024prism} and \textbf{GOOD}~\citep{wang-etal-2019-persuasion}. Both contain explicit personal value profiles and user behaviors across diverse scenarios. % , enabling the measurement of user preferences and context-dependent value trade-offs.
% PRISM consists of survey-style personal value profiles and turn-level user-model interactions, where each has an open-ended question and preference feedback to candidate responses. GOOD is constructed from the PersuasionForGood corpus~\citep{wang-etal-2019-persuasion}, comprising persuasion dialogues with rich participant value annotations.
% After preprocessing and filtering, PRISM and GOOD contain 2{,}131 and 2{,}696 instances respectively. We adopt an 80/20 split for training and testing on both datasets.
% Each PRISM instance comprises a user profile, a question, a preferred answer and a dispreferred answer. Each GOOD instance contains a user profile, a question and a user-generated response.

\paragraph{Datasets}
We evaluate BaCVA on \textbf{PRISM}~\citep{kirk2024prism} and \textbf{GOOD}~\citep{wang-etal-2019-persuasion}, both of which provide personal value profiles and user behaviors across diverse scenarios. 
PRISM contains open-ended questions with preferred/dispreferred candidate responses, while GOOD contains persuasion dialogues with value annotations and user-generated responses. 
After preprocessing, they contain $2,131$ and $2,696$ instances, respectively, and are split 80/20 for training and testing. Additional results on other datasets are provided in Appendix~\ref{app:hy_exp}.

\paragraph{Evaluation Metrics} As formalized in Sec.~\ref{subsec:task_definition}, we represent value orientation as a multi-dimensional value vector. First, we report \textbf{Mean Absolute Error (MAE)} between the predicted user value preferences and the value of preferred answers. Second, we compute \textbf{Spearman’s rank correlation coefficient (Correlation)} to measure the consistency of value priorities. Third, we report the \textbf{Accuracy} that the preferred responses are ranked higher than dispreferred responses.
Detailed dataset construction, statistics, and metric definitions are provided in Appendix~\ref{app:exp_setup}.

\paragraph{Baselines}
We compare BaCVA with baselines spanning non-personalized and inference-time personalized alignment. 
\textbf{DirectAnswer} directly generates responses using the generally aligned backbone model. 
For personalized value alignment, we mainly consider inference-time approaches that incorporate static user value profiles to steer answer generation, including four groups: (1) \textbf{Value Prompt}~\citep{kang2024causal_align} prompts the value profile to steer outputs, \textbf{COUPLE}~\citep{guo2026counterfactual} conducts counterfactual reasoning, and \textbf{MetaAligner}~\citep{yang2024metaaligner} rewrites model outputs to align them with target values; (2) \textbf{PAD}~\citep{chen2024pad} and \textbf{MOD}~\citep{shi2024decoding} learn value-aware reward models to steer decoding-time token distribution towards personal values; (3) \textbf{ValuesRAG}~\citep{seo2025valuesrag} retrieves historical data under similar scenarios and value profiles as example for decision-making; and (4) \textbf{PersonValue} is implemented by us that directly compares the static value profiles with values extracted from the preferred respones. To further compare BaCVA with training-time alignment approaches, we additionally evaluate RLHF~\citep{christiano2017deep} and MORLHF~\citep{li2020deep} under the same backbone, data splits, and input information, with full implementation details and results provided in Appendix~\ref{app:training_time_baselines}.

% \paragraph{Baselines}
% We compare BaCVA with diverse representative baselines covering both non-personalized and personalized paradigms. 
% For non-personalized alignment, we include a baseline \textbf{DirectAnswer} that directly generates answers to prompts using the generally aligned backbone model.
% For personalized value alignment, we mainly consider inference-time approaches that incorporate static user value profiles to steer answer generation. Four groups of baselines are included: (1) \textbf{Value Prompt}~\citep{kang2024causal_align} directly prompts the value profile to steer outputs, \textbf{COUPLE}~\citep{guo2025counterfactual} conducts counterfactual reasoning, and \textbf{MetaAligner}~\citep{yang2024metaaligner} rewrites model outputs to better align them with target values; (2) \textbf{PAD}~\citep{zhu2024personality} and \textbf{MOD}~\citep{shi2024decoding} learn value-aware reward models to steer decoding-time token distribution towards personal values; (3) \textbf{ValuesRAG}~\citep{seo2025valuesrag} retrieves historical data with similar scenarios and value profiles as example for decision-making; and (4) \textbf{PersonValue} is implemented by us that directly compares the static value profiles with values extracted from the preferred respones.

\textbf{BaCVA} dynamically estimates personalized values across varying contexts via posterior inference. % , thus we also compare the predicted values with that of the chosen responses.

% We compare BaCVA with a diverse set of representative baselines, covering both non-personalized and personalized value alignment paradigms. The compared methods are organized according to their personalization capability and output type.
% \textbf{Universal alignment methods} ignore user-specific value profiles and operate solely on the input question or context $Q$. We include a question-only predictor (Q-Only) and a direct answer generation baseline (DirectGen) as representative baselines, both yielding universal, user-agnostic outputs.
% \textbf{Personalized alignment methods} incorporate user value profiles $V_P$, either alone or jointly with the question $Q$. 
% Among them, \emph{prediction-based} approaches estimate value ratings or preference scores for candidate responses, including a profile-only predictor ($VA\equiv VP$) and a retrieval-augmented method (\emph{ValuesRAG}~\citep{seo2025valuesrag}).
% \emph{Answer-based} personalized methods directly generate or select responses conditioned on value information, covering \emph{Value Prompt}~\citep{kang2024causal_align}, \emph{MetaAligner}~\citep{yang2024metaaligner}, \emph{COUPLE}~\citep{guo2025counterfactual}, \emph{PAD}~\citep{zhu2024personality}, and \emph{MOD}~\citep{shi2024decoding}.
% Our method, \emph{BaCVA}, also belongs to this category but differs by performing explicit Bayesian posterior inference and adaptive prior weighting at inference time.

\paragraph{Implementation details.}
We use Qwen3-8B as the backbone model across all comparable methods and control input as only personal value profile and question to ensure fair comparisons. The model for value extraction in both our model and evaluation is GPT-5-nano, whose consistency with humans has been verified, achieving a score $> 90$. Unless otherwise specified, we use the 10 Schwartz Basic value dimensions to represent value profiles in the main experiments. Additional experiments on other value systems to validate the robustness are provided in Appendix~\ref{app:hy_exp}.
All models are trained using the Adam optimizer.
All experiments are conducted using NVIDIA A100 GPUs. Full implementation details, hyperparameters, and more experimental settings are in Appendix~\ref{app:exp_setup}. 

\subsection{Personalized Alignment Performance}
Tab.~\ref{tab:main_prism_good} presents the results comparing BaCVA with baselines on personalized value alignment across PRISM and GOOD, with three key findings.
% which reveal three key findings.

\textit{First, incorporating personalization is crucial for LLM value alignment}. Across both benchmarks, personalized alignment methods such as Value Prompt, MOD and ValuesRAG substantially outperform the non-personalized baseline DirectAnswer. This observation demonstrates that personal value-agnostic alignment fails to capture the diversity of individual preferences.
\textit{Second, static personal value profiles lead to sub-optimal personalized alignment.} In particular, the PersonValue baseline, which enforces adherence to static personal values across varying prompts, achieves merely 56.56\% accuracy on PRISM. This highlights a fundamental limitation that static personal values cannot faithfully represent users’ context-dependent preferences across diverse scenarios.
\textit{Third, BaCVA, which explicitly integrates personal value profiles with scenario-specific value salience and optimizes toward posterior personalized preferences, achieves the best personalization result across both benchmarks.} On \textsc{PRISM}, BaCVA attains the lowest MAE and the highest accuracy, while on \textsc{GOOD} it significantly improves both MAE and correlation. These consistent gains strongly validate our motivation in this paper and the effectiveness of our proposed method. Additionally, BaCVA also consistently outperforms the training-time alignment baselines reported in Appendix~\ref{app:training_time_baselines}.

\begin{figure}[t]
    \centering
        \includegraphics[width=\linewidth]{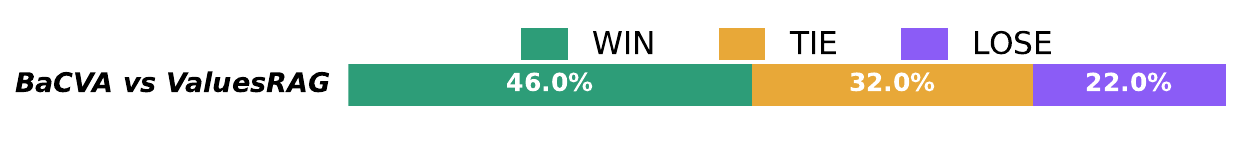}
        \caption{Performance of human evaluation}
    \label{fig:human}
\end{figure}

\paragraph{Human Evaluation}
Beyond automatic evaluation, we conduct a human study on 50 samples corresponding to 50 distinct user profiles, with each sample independently assessed by two evaluators, who compare how closely the contextual personalized values predicted by BaCVA and ValuesRAG mirror the values of the preferred answer. As shown in Fig.~\ref{fig:human}, BaCVA wins in 46.0\% and loses in only 22.0\% of cases, outperforming ValuesRAG. Details are provided in Appendix~\ref{app:human_eval}.

% Table~\ref{tab:main_prism_good} reveals three clear trends.
% First, inference-time personalized alignment methods consistently outperform universal baselines that ignore user value profiles. For instance, on \textsc{PRISM}, the universal predictor Q-Only performs poorly (MAE 3.519), while personalized methods substantially reduce prediction error and yield more stable results. Similar patterns are observed on \textsc{GOOD}, where universal baselines exhibit weak or even negative correlations.

% Second, methods that explicitly analyze user value information rather than relying on shallow conditioning tend to achieve better performance. ValuesRAG exemplifies this trend, significantly improving over prompt-based personalized methods and demonstrating that retrieving and reasoning over user-related value evidence is an effective strategy for personalized alignment.

% Third, by deeply integrating user value preferences with universal task semantics through Bayesian modeling, BaCVA achieves the best overall performance across both benchmarks. On \textsc{PRISM}, BaCVA attains the lowest MAE and the highest Recall@1, while on \textsc{GOOD} it substantially improves both error and correlation. These consistent gains indicate that explicitly modeling the interaction between value preferences and task context is crucial for robust and accurate personalized alignment.

\begin{table}[t]
\centering
\small
\renewcommand{\arraystretch}{1.4}
\setlength{\tabcolsep}{4pt}
\resizebox{1.0\linewidth}{!}{
\begin{tabular}{lccccc} % {l *{5}{>{\centering\arraybackslash}X}}
\toprule
 & \multicolumn{3}{c}{\textbf{PRISM}} & \multicolumn{2}{c}{\textbf{GOOD}} \\
\cline{2-4} \cline{5-6}
\textbf{Method} 
& \textbf{MAE}~$\downarrow$ 
& \textbf{Corr}~$\uparrow$ 
& \textbf{Accuracy}~$\uparrow$
& \textbf{MAE}~$\downarrow$
& \textbf{Corr}~$\uparrow$ \\
\midrule

\textbf{BaCVA}
& \textbf{0.728} & \textbf{0.700} & \textbf{81.72} 
& \textbf{1.118} & \textbf{0.769} \\

\hline
% No Value Prior & 0.936 & 0.598 & 71.54 & 1.483 & 0.705 \\

% Single Prior ($V_P$)
~ w/o Scenario-View
& 0.819 & 0.645 & 76.32
& 1.325 & 0.724 \\

% Single Prior ($V_U$)
~ w/o Person-View
& 0.798 &  0.607 & 77.06
& 1.297 & 0.730 \\

% BaCVA w/o Gate
~ w/o Gated Fusion
& 0.762 & 0.671 & 79.44 
& 1.163 & 0.754 \\

% No Bayesian Inference
~ w/o Prior Constraint
& 0.800 & 0.639 & 78.18
& 1.203 & 0.746 \\ 
\bottomrule
\end{tabular}
}
\caption{Ablation study on \textsc{PRISM} and \textsc{GOOD}.}
\label{tab:ablation_prism_good}
\end{table}

% ~ w/o KL Divergence
% & 0.751 & 0.688 & 80.05
% % & 1.141 & 0.762 \\ 

\subsection{Ablation Study}
To systematically investigate the contribution of each component in BaCVA, we conduct an ablation study by removing or modifying key modules. (1) \textbf{w/o Scenario-View} and \textbf{w/o Person-View} approximate the posterior value preferences from only one of the dual perspectives. (2) \textbf{w/o Gated Fusion} replaces the adaptive gating mechanism with an averaged fusion of the two views across all contexts. (3) \textbf{w/o Prior Constraint} removes the KL divergence terms in Eq.(\ref{eq:dual_loss}). Tab.~\ref{tab:ablation_prism_good} reports the ablation results on \textsc{PRISM} and \textsc{GOOD}.  % , akin to learning from observed data without prior information

% by selectively removing or modifying key modules, including value priors, Bayesian posterior inference, and the adaptive gating mechanism. \textbf{No Value Prior} removes both the personal value prior $V_P$ and the universal value prior $V_U$, forcing the model to rely solely on contextual representations.
% \textbf{Single Prior ($V_P$)} and \textbf{Single Prior ($V_U$)} retain only one static value prior while disabling the other, without any contextual adaptation.
% \textbf{No Bayesian Inference} removes the Bayesian posterior inference module and directly fuses priors and contextual features in a deterministic manner.
% \textbf{BaCVA w/o Gate} disables the adaptive gating mechanism and assigns fixed weights to different priors across all contexts.
% The full \textbf{BaCVA} model incorporates all components.

\textbf{Results Analysis}
Using merely a single view (i.e., \emph{w/o Scenario-View} and \emph{w/o Person-View}) consistently results in degraded performance on both benchmarks. This confirms that both views capture effective and complementary aspects for personalized alignment.
Besides, removing the adaptive gating mechanism (\emph{w/o Gated Fusion}) causes a performance drop. This demonstrates that the interaction between the two views is highly context-dependent rather than simply additive. The gated fusion is essential to distinguish their relative importance across different scenarios.
Finally, the prior constraints of both personal values and scenario-specific value salience are also critical for personalized alignment by avoiding collapsed prediction. This also validates that both views provide meaningful prior knowledge.

\begin{figure}[t]
    \centering
    \includegraphics[width=1.0\linewidth]{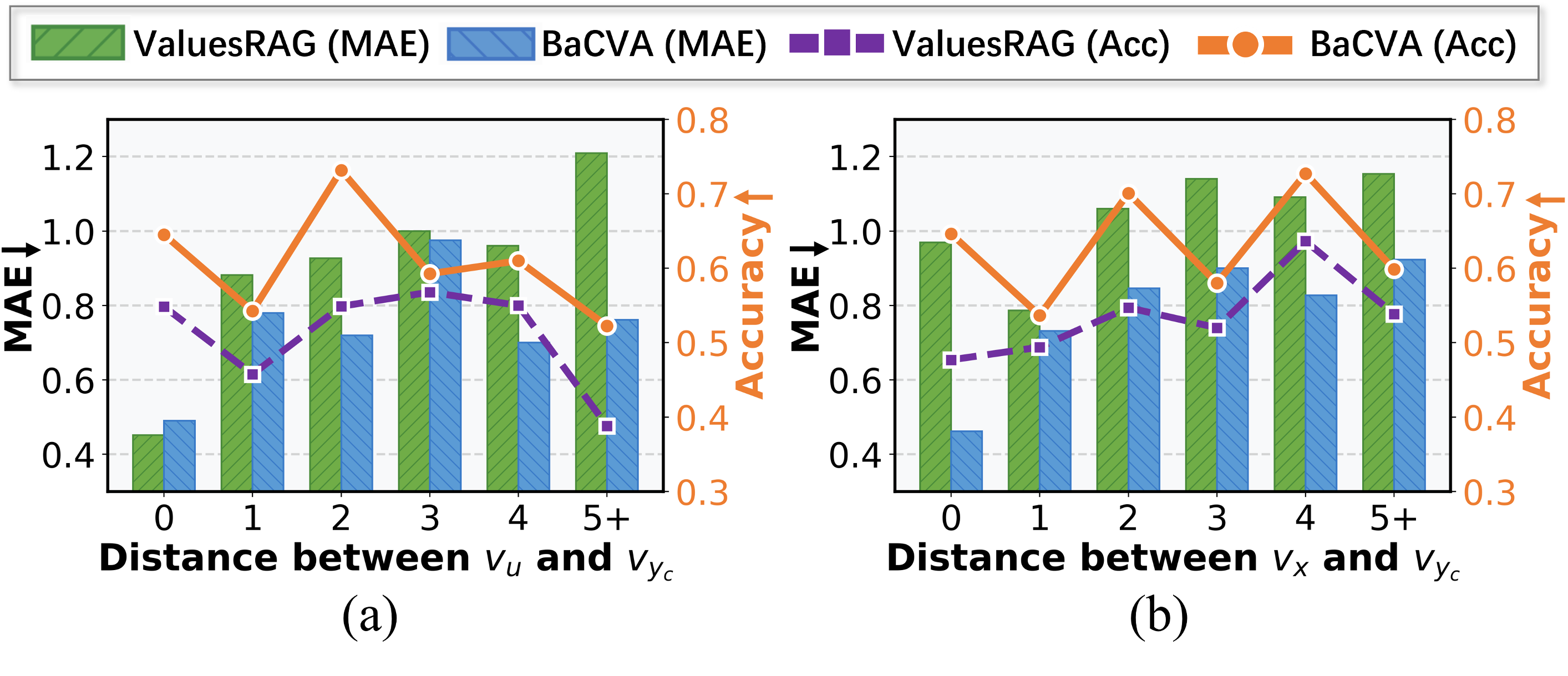}
    % \begin{subfigure}{0.48\linewidth}
    %     \centering
    %     \includegraphics[width=\linewidth]{figures/exp_controllability_1_prism.png}
    %     \caption{Effect of $V_P - V_A$ Distance}
    % \end{subfigure}
    % \hfill
    % \begin{subfigure}{0.48\linewidth}
    %     \centering
    %     \includegraphics[width=\linewidth]{figures/exp_controllability_2_prism.png}
    %     \caption{Effect of $V_{\text{uni}} - V_A$ Distance}
    % \end{subfigure}
    \caption{Fine-grained personalization on \textsc{PRISM}.}
    \label{fig:controllability}
\end{figure}

% \begin{figure}[t]
%     \centering
%     \begin{subfigure}{0.48\linewidth}
%         \centering
%         \includegraphics[width=\linewidth]{figures/exp_efficiency_1.png}
%         \caption{Training convergence on \textsc{PRISM}}
%     \end{subfigure}
%     \hfill
%     \begin{subfigure}{0.48\linewidth}
%         \centering
%         \includegraphics[width=\linewidth]{figures/exp_efficiency_2.png}
%         \caption{Training convergence on \textsc{GOOD}}
%     \end{subfigure}

%     \caption{Training efficiency analysis across datasets.}
%     \label{fig:efficiency}
% \end{figure}

\begin{figure}[t]
    \centering
\includegraphics[width=\linewidth]{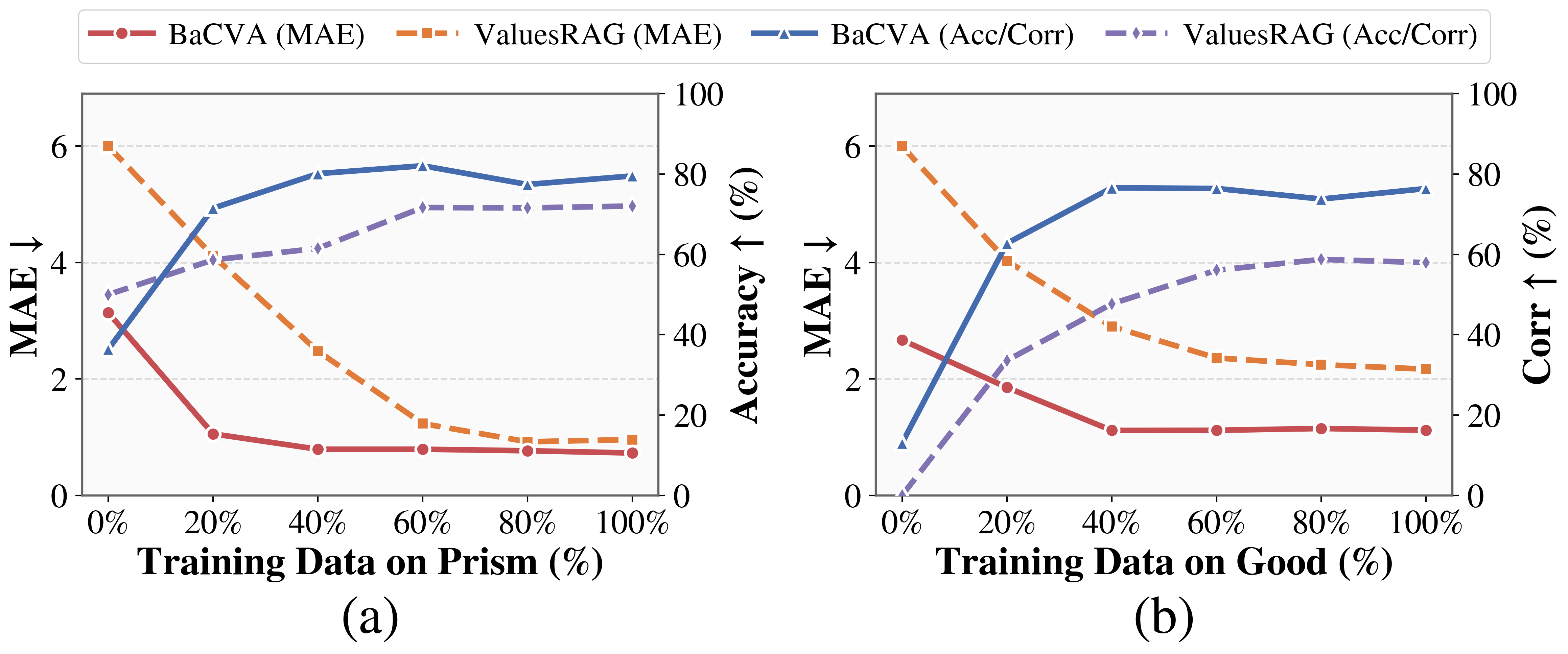}
    \caption{Training efficiency analysis across datasets.}
    \label{fig:efficiency}
\end{figure}

% \begin{figure}[t]
%     \centering
%     \begin{subfigure}{0.48\linewidth}
%         \centering
%         \includegraphics[width=\linewidth]{figures/exp_gate_1_prism.png}
%         \caption{}
%     \end{subfigure}
%     \hfill
%     \begin{subfigure}{0.48\linewidth}
%         \centering
%         \includegraphics[width=\linewidth]{figures/exp_gate_2_prism.png}
%         \caption{}
%     \end{subfigure}

%     \caption{Gate effect and alignment behavior under distributional scenarios.}
%     \label{fig:gate}
% \end{figure}

\begin{figure}[t]
    \centering
    
    \includegraphics[width=\linewidth]{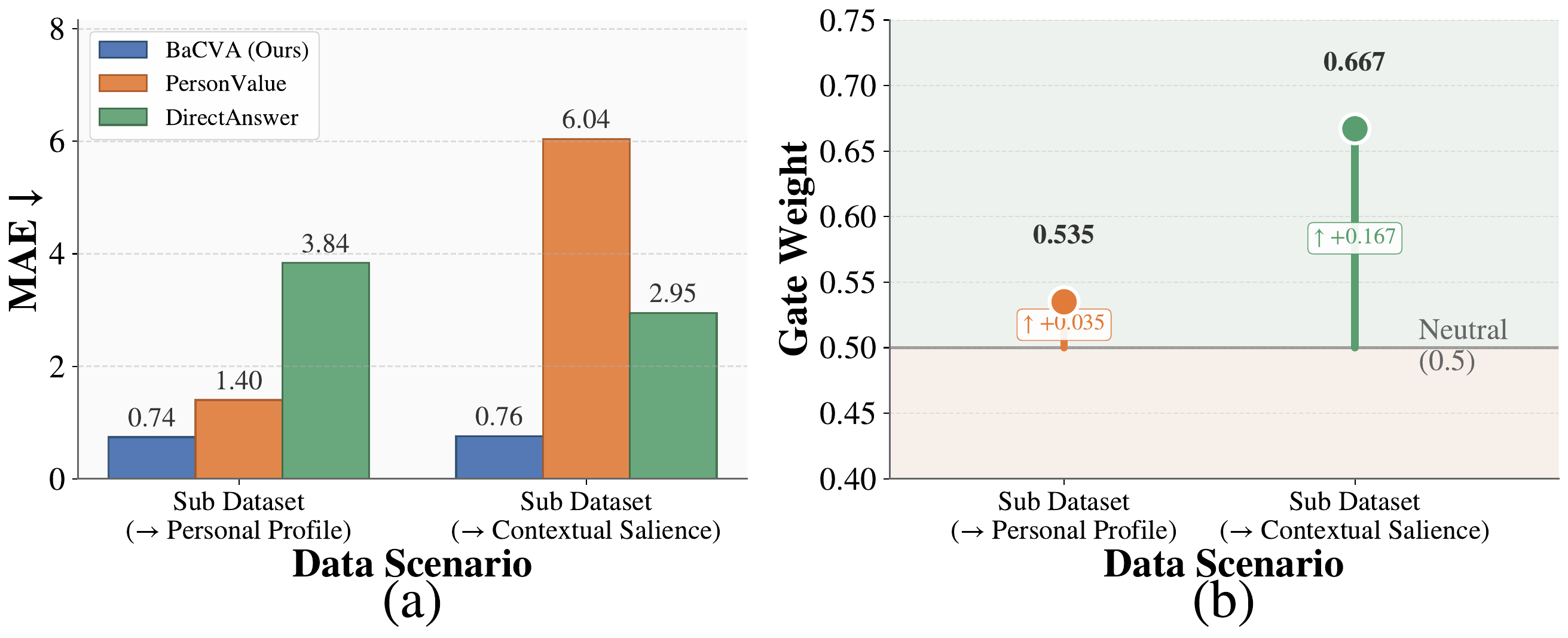}
    \caption{Effects of the gated fusion on PRISM.}
    % \caption{Gate effect and alignment behavior under distributional scenarios.}
    \label{fig:gate}
\end{figure}

\subsection{Analysis Experiments}
\label{exp:analysis}

% With both personal value profiles and scenario-specific value salience as priors, BaCVA captures finer-grained deviation from these priors to accurately infer posterior preferences. To access this adaptability, we examine personalization performance on test samples grounded by the distances between the values of the preferred response $v_{y_c}$ and (i) the personal prior $v_{u}$ and (ii) the scenario prior $v_{x}$. Larger distances correspond to cases where the final preferences substantially deviate from the prior. Fig.~\ref{fig:controllability} shows the results of BaCVA and a strong baseline ValuesRAG.

% In Fig.~\ref{fig:controllability} (a), we observe that ValuesRAG performs well only on low-distance samples and degrades sharply as the distance increases, indicating a strong dependence on static personal value priors. In contrast, BaCVA consistently achieves better results across all distance intervals. As for Fig.~\ref{fig:controllability} (b), BaCVA maintains stable performance regardless of how closely the preferred response originally aligns with the scenario value salience, while ValuesRAG shows no superiority on any groups.
% These results confirm that BaCVA has fine-grained adaptability to infer posterior personalized preference across contexts, rather than just adhering to personal value or scenario value priors.

\paragraph{Fine-grained Adaptability Analysis}
We examine whether BaCVA captures fine-grained shifts from priors by grouping test samples according to the distance between the true preferred response value $v_{y_c}$ and two priors: the personal prior $v_u$ and the scenario prior $v_x$. A larger distance means that the final contextual preference deviates more from the corresponding prior, making the sample more challenging for conventional methods that simply follow the prior. 
As shown in Fig.~\ref{fig:controllability}(a), ValuesRAG performs well only when $v_{y_c}$ is close to $v_u$, but degrades sharply as the distance increases, indicating its strong dependence on static personal values. In contrast, BaCVA remains stable across different distances from both priors. These results suggest that BaCVA can capture nuanced preference shifts to infer posterior preferences even when the context heavily conflicts with priors. Moreover, a complementary conflict-level analysis further shows that BaCVA reduces MAE by 17.8\% relative to ValuesRAG on the high-conflict subset, where answer-level values deviate substantially from the global user profile (Appendix~\ref{app:conflict_analysis}).
% infer context-dependent posterior preferences beyond either personal or scenario priors.

% \textcolor{blue}{To further verify domain-level robustness, we additionally perform a cluster-level analysis on \textsc{PRISM} in Tab.~\ref{tab:cluster_analysis}. The results show that BaCVA consistently reduces MAE across five most common semantic clusters, including social values, life advice, politics/civic issues, religion/tradition, and relationships. This indicates that BaCVA's fine-grained adaptation is not limited to aggregate performance, but remains stable across diverse contextual domains. We also provide additional evaluation on AlignX, a dataset with more diverse and shifting contextual scenarios, in Appendix~\ref{app:addition_exp}.
% }

A cluster-level analysis on \textsc{PRISM} further confirms BaCVA's domain-level robustness, with consistent MAE reductions across the five largest semantic clusters (Appendix~\ref{appendix:cluster}).

\paragraph{Training Data Efficiency}
We evaluate data efficiency by training BaCVA with different proportions of PRISM and GOOD data. As shown in Fig.~\ref{fig:efficiency}, BaCVA achieves strong performance with limited supervision and improves steadily as more data is used, whereas ValuesRAG converges more slowly and to a weaker final result. This indicates that BaCVA can learn effective value-alignment patterns in low-data regimes.

% Although RAG-based methods generally require less training data, the inherent complexity of value-centric scenarios
% still necessitates a substantial amount of data to achieve reliable performance.
% This indicates that BaCVA is data-efficient and capable of learning effective value alignment patterns under limited
% supervision.

\begin{figure}[t]
    \centering
\includegraphics[width=1.0\linewidth]{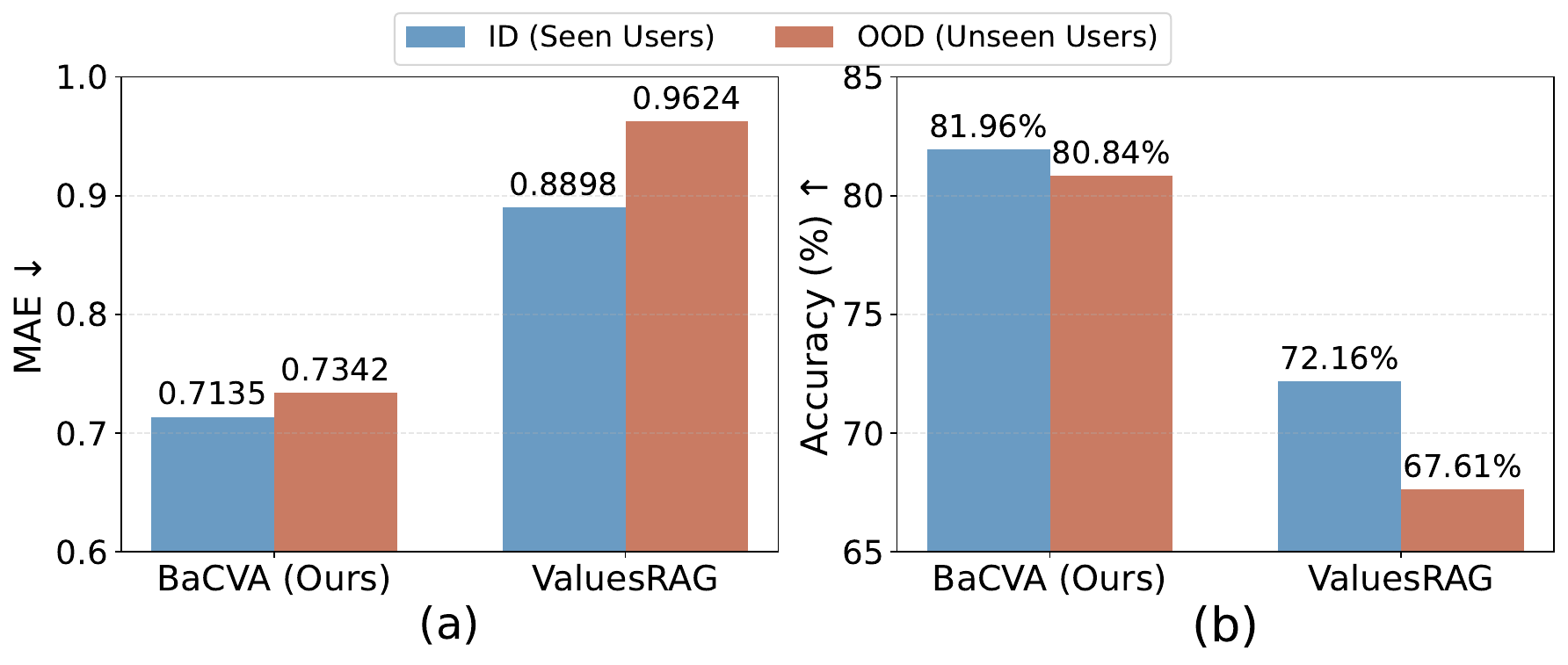}
    \caption{Generalizability evaluation on \textsc{PRISM}.}
    \label{fig:ood}
\end{figure}

\paragraph{Generalization to Cold-Start Users}
We further evaluate generalization to unseen users by splitting the PRISM test set into seen users and 20\% cold-start users absent from training. As shown in Fig.~\ref{fig:ood}, BaCVA shows a much smaller performance drop than ValuesRAG, demonstrating stronger robustness to unseen users.
% We investigate the generalization ability of BaCVA across users. Specifically, we hold out 20\% of users from the \textsc{PRISM} dataset as out-of-domain (OOD) users and evaluate the models on these unseen users. As shown in Fig.~\ref{fig:ood}, BaCVA consistently outperforms ValuesRAG on OOD users, indicating stronger user-level generalization and better robustness to distribution shifts.

% \paragraph{Gated Fusion Analysis.}
% BaCVA infers posterior values from two complementary views and integrates them via a gated fusion, which analyzes the contexts to emphasize the more effective view. To validate the effectivenesss of this gated fusion, we examine the model performance on two subsets: (i) \emph{personal value-dominant cases} where the values of preferred response is closer to the user's personal values (i.e., $|v_{y_c} - v_u| < |v_{y_c} - v_{\text{x}}|$), and (ii) \emph{scenario-dominant cases}. % Besides, we also visualize the gated weights assigned to the corresponding view in the two groups. 

% As shown in Fig.~\ref{fig:gate}(a), PersonValue performs better on the personal value-dominant subset while DirectAnswer is better on scenario-dominant cases. Consistently, BaCVA achieves the best performance across both subsets, indicating the effects of complementary views. Furthermore, we visualize the gated weights assigned to the corresponding view on the two groups. As expected, the gating mechanism emphasizes the correct view. For example, in the scenario-dominant subset, the scenario view is assigned a much larger weight.

\paragraph{Gated Fusion Analysis.}
% BaCVA integrates the person-view and scenario-view through a context-aware gate. 
% To examine whether the gate selects the more informative view, we split PRISM into \emph{personal value-dominant} cases, where $v_{y_c}$ is closer to $v_u$ than to $v_x$, and \emph{scenario-dominant} cases. 
% As shown in Fig.~\ref{fig:gate}(a), PersonValue performs better in personal value-dominant cases, while DirectAnswer performs better in scenario-dominant cases. 
% BaCVA achieves the best performance on both subsets, and Fig.~\ref{fig:gate}(b) further shows that the gate assigns larger weights to the corresponding dominant view.

BaCVA fuses the person-view and scenario-view through a context-aware gate. To verify whether the gate correctly identifies the more informative view across varying contexts, we split PRISM into personal-dominant cases, where $v_{y_c}$ is closer to $v_u$ than to $v_x$, and scenario-dominant cases. Fig.~\ref{fig:gate}(a) shows that PersonValue performs competitively in personal-dominant settings but fails catastrophically when situational shifts occur, whereas \emph{DirectAnswer} is opposite. BaCVA consistently performs best on both. Fig.~\ref{fig:gate}(b) further shows that the gate assigns larger weights to the corresponding dominant view.
% Fig.~\ref{fig:gate}(a) shows that PersonValue and DirectAnswer are respectively stronger in these two subsets, while BaCVA performs best on both. Fig.~\ref{fig:gate}(b) further shows that the gate assigns larger weights to the corresponding dominant view.

\paragraph{Cross-Extractor Robustness}
Cross-extractor evaluation using independently constructed Gemini-3-Flash and Qwen3-235B-A22B test targets confirms that BaCVA's gains are not tied to GPT-5-nano, yielding 21.2\% lower MAE and 9.56 percentage points higher accuracy than ValuesRAG on average (Appendix~\ref{app:cross_extractor}).

\subsection{Hyper-parameter Analysis}
\label{exp:hyper}

We provide a sensitivity analysis of the KL-divergence weight $\lambda_1$ in Appendix~\ref{app:hy_exp}. 
Besides, we further examine the robustness of BaCVA under different backbone LLMs as the value salience estimator, prompt designs, and value representations. Specifically, we evaluate BaCVA with value profiles represented by the 5-dim Moral Foundations Theory and the 301-dim value system from Daily-Dilemma. 
The consistent gains across these settings show that BaCVA is robust to both implementation choices and value-system variations. Meanwhile, to evaluate BaCVA's generalizability across backbone models, we further implement it with Llama3-8B under the same experimental setting, where it reduces MAE by 37.9\% and improves accuracy by 19.06 percentage points over ValuesRAG in Appendix~\ref{app:backbone_robustness}. To evaluate robustness across scenario-prior estimators, we compare GPT-5-nano, Gemini-3-Flash, and Qwen3-235B-A22B, obtaining an average within-one agreement of 89.15\% and a standard deviation of only 0.027 in downstream PRISM MAE in Appendix~\ref{app:prior_estimator_robustness}. To assess whether BaCVA training degrades general-purpose capabilities, we evaluate the original backbone and BaCVA on four standard benchmarks, obtaining comparable average performance of 77.69\% and 78.74\%, respectively (Appendix~\ref{app:general_capability}).

\subsection{Downstream Application}
\label{exp:downstream}

Furthermore, we evaluate the effectiveness of our inferred posterior values in guiding downstream personalized response generation using the PRISM dataset. We implement two variants for generation: (1) \textbf{BaCVA (Two-stage)} applies the posterior personalized values estimated by BaCVA as the input and trains a downstream response generator. (2) \textbf{BaCVA-E2E} jointly optimizes the BaCVA for posterior value inference with subsequent response generator. Here, MAE, Spearman correlation, and pairwise ranking accuracy are computed under the answer-level protocol detailed in Appendix~\ref{appendix:e2e}. As shown in Tab.~\ref{tab:e2e_main}, BaCVA-E2E achieves the best overall performance, slightly outperforming the two-stage BaCVA.

% We evaluate downstream personalized response generation on PRISM under an answer-level protocol with flexible active value dimensions per instance. 
% For comparison, \textbf{BaCVA (Two-stage)} first uses the trained BaCVA value inferrer to estimate contextualized value vectors, which are then used as fixed guidance for training the downstream response generator. 
% In contrast, \textbf{BaCVA-E2E} jointly optimizes contextual value inference and response generation. 
% Accordingly, MAE and Spearman correlation are computed over answer-specific active dimensions, while pairwise ranking accuracy follows the same setting (Appendix~\ref{appendix:e2e}). 
% As shown in Tab.~\ref{tab:e2e_main}, BaCVA-E2E achieves the best overall performance, slightly outperforming the two-stage BaCVA.

Moreover, qualitative cases in Appendix~\ref{appendix:e2e} further show that BaCVA-E2E can generate coherent value-grounded responses, whereas ValuePrompt often fails.

% \textcolor{blue}{Moreover, Fig.~\ref{fig:case_study_2} illustrates a representative case where ValuePrompt, despite access to the true value profile, merely lists values by name, whereas BaCVA-E2E produces a coherent, value-grounded response. Implementation details and more cases are in Appendix~\ref{appendix:e2e}.}

\begin{table}[t]
\centering
 % Main results on the PRISM test set using answer comparison. \textbf{Bold} indicates the best performance per column.
\resizebox{\columnwidth}{!}{%
\begin{tabular}{lccc}
\toprule
\textbf{Method} & \textbf{MAE} $\downarrow$ & \textbf{Acc} $\uparrow$ & \textbf{Spearman} $\uparrow$ \\
\midrule
DirectAnswer              & 1.1794 & 0.3226 & 0.2483 \\
ValuePrompt               & 0.9897 & 0.4892 & 0.3121 \\
COUPLE                    & 0.5596 & 0.6434 & 0.4432 \\
MetaAligner               & 0.7787 & 0.5125 & 0.4010 \\
PAD                       & 0.6444 & 0.5341 & 0.4463 \\
MOD                       & 0.6707 & 0.5448 & 0.4013 \\
ValuesRAG                 & 0.6229 & 0.5529 & 0.4596 \\
\midrule
BaCVA (Two-stage)         & 0.3835 & 0.7975 & 0.4782 \\
\textbf{BaCVA-E2E}        & \textbf{0.3776} & \textbf{0.7996} & \textbf{0.4812} \\
\bottomrule
\end{tabular}%
}
\caption{Results of downstream personalized response generation on PRISM. \textbf{Bold} indicates the best results.}
\label{tab:e2e_main}
\end{table}

\section{Conclusion}
% Motivated by the limitation that current personalized value alignment relies on static personal values and overlooks context-dependent value salience, we propose BaCVA, an inference-time Bayesian framework that regards both stable personal values and context-specific value salience as priors, then infers posterior personalized values. Experiments demonstrated that BaCVA achieves more accurate and adaptive personalization, as well as improved data efficiency and generalization.

To overcome the limitation that current personalized value alignment relies on static user value profiles, BaCVA treats personal values and context-specific value salience as priors and infers posterior personalized values at inference time. Experiments validate improved personalization accuracy, adaptability, data efficiency, and generalization.

% In realistic deployment settings, personalized value alignment must address both inter-user heterogeneity and intra-user, context-dependent trade-offs. We propose a Bayesian framework that models stable user preferences as a prior and infers scenario-specific value orientations via context-conditioned posterior updates, improving alignment quality and interpretability over fine-tuning and prompting baselines. Future work includes continual personalization, robustness to sparse or noisy feedback, and privacy-preserving deployment.
\section*{Limitations}

Although BaCVA achieves promising performance on personalized value alignment across two benchmarks, this work still has several limitations. 

\emph{First}, the evaluation method may introduce noise, as it relies on user preferences annotated over a limited set of candidate responses, which may not fully capture the faithful user values but only compromised preferences. Moreover, our experiments are primarily conducted on offline datasets; although human evaluation is included, the approach lacks validation through large-scale real-world or online deployment. \emph{Second}, the cross-dataset generalization of our model remains an important aspect to be further validated. \emph{Third}, the datasets we used for evaluation contain users' self-reported value profiles, which may be noisy as users may consciously or unconsciously curate, simplify, or embellish their profiles. 
\emph{Last}, the estimator remains an aligned-LLM approximation of population-level value salience, which future work may improve with more diverse models, alignment strategies, and cultural assumptions.

In future work, we plan to address these limitations by exploring more realistic and comprehensive evaluation settings, investigating cross-dataset generalization, and developing more robust mechanisms to model and mitigate noise in user-provided information.

% Although BaCVA achieves personalized alignment by integrating personal values with context-dependent value salience at inference time and shows promising performance on benchmark datasets, this work still has several limitations. First, the evaluation methodology may introduce noise, as it relies on user preferences annotated over a limited set of candidate responses, which may not fully capture the diverse dimensions of user values and preferences. Moreover, our experiments are primarily conducted on offline datasets; although human evaluation is included, the approach lacks validation through large-scale real-world or online deployment. Second, the cross-dataset generalization and transferability of our model remain an important aspect to be further validated, as our current study focuses on in-dataset evaluation. Third, users’ self-reported information may be inherently noisy, as users may consciously or unconsciously curate, simplify, or embellish their profiles. In future work, we plan to address these limitations by exploring more realistic and comprehensive evaluation settings, investigating cross-dataset generalization, and developing more robust mechanisms to model and mitigate noise in user-provided information.

\section*{Ethical Considerations}

This work aims to improve personalized value alignment in large language models; however, the inferred “value preferences” reflect latent patterns captured from observed interactions and annotations, and should be viewed as approximate rather than complete representations of human values. It should not be used to directly judge individuals or replace human ethical decision-making. Personalized value modeling may also involve privacy and bias risks. Therefore, any real-world deployment should strictly follow data protection regulations, avoid manipulative or unethical uses, and be continuously evaluated within a responsible AI framework.

\section*{Acknowledgments}

This work was supported by the Beijing Nova Program (Grant No.~202604841294).

% \section*{Acknowledgments}

% This document has been adapted
% by Steven Bethard, Ryan Cotterell and Rui Yan
% from the instructions for earlier ACL and NAACL proceedings, including those for
% ACL 2019 by Douwe Kiela and Ivan Vuli\'{c},
% NAACL 2019 by Stephanie Lukin and Alla Roskovskaya,
% ACL 2018 by Shay Cohen, Kevin Gimpel, and Wei Lu,
% NAACL 2018 by Margaret Mitchell and Stephanie Lukin,
% Bib\TeX{} suggestions for (NA)ACL 2017/2018 from Jason Eisner,
% ACL 2017 by Dan Gildea and Min-Yen Kan,
% NAACL 2017 by Margaret Mitchell,
% ACL 2012 by Maggie Li and Michael White,
% ACL 2010 by Jing-Shin Chang and Philipp Koehn,
% ACL 2008 by Johanna D. Moore, Simone Teufel, James Allan, and Sadaoki Furui,
% ACL 2005 by Hwee Tou Ng and Kemal Oflazer,
% ACL 2002 by Eugene Charniak and Dekang Lin,
% and earlier ACL and EACL formats written by several people, including
% John Chen, Henry S. Thompson and Donald Walker.
% Additional elements were taken from the formatting instructions of the \emph{International Joint Conference on Artificial Intelligence} and the \emph{Conference on Computer Vision and Pattern Recognition}.

\bibliography{custom}

\clearpage
\appendix

\section{Method Details}
\label{app:method}

\subsection{Generic ELBO for Personalized Value Inference}
\label{app:generic_elbo}

\paragraph{Setup}
For a user $u$ and a question (scenario) $x$, let $\mathbf v_u$ denote the user-level value profile, and let $\mathbf v_{u,x}$ denote the latent question-specific personalized values.
Given a preference triple $(x,y_c,y_r)$, we denote the observed preference event as
\begin{equation}
o := (y_c \succ y_r),
\end{equation}
i.e., response $y_c$ is preferred over $y_r$ under scenario $x$.

\paragraph{Bayesian formulation}
We define a prior conditioned on both the user and scenario:
\begin{equation}
p_\theta(\mathbf v_{u,x}\mid \mathbf v_u, x),
\end{equation}
and a preference likelihood
\begin{equation}
p_\theta(o \mid \mathbf v_{u,x}, x)
\;\;=\;\;
p_\theta(y_c \succ y_r \mid \mathbf v_{u,x}, x).
\end{equation}
By Bayes' rule, the posterior is
\begin{align}
p_{\theta}(\mathbf v_{u,x}\mid \mathbf v_u, x, o) =
\frac{p_\theta(o\mid \mathbf v_{u,x})\, p_\theta(\mathbf v_{u,x}\mid \mathbf v_u, x)}{p_{\theta}(o)},
\end{align}
which matches Eq.~(\ref{eq:posterior}) in the main paper.

\paragraph{Marginal likelihood}
% Given the Bayesian formulation in Eq.~(\ref{eq:posterior}), learning can be posed as maximizing the conditional evidence (marginal likelihood) of the observed preference event
The likelihood $o := (y_c \succ y_r)$ marginalizes over the latent personalized values $\mathbf v_{u,x}$:
\begin{align}\label{eq:marginal_basic}
    p_{\theta}(o) & = \int p_{\theta}(o \mid \mathbf v_{u,x}, x) \\
    \times
    & p_{\theta}(\mathbf v_{u,x} \mid \mathbf v_u, x)
\, d\mathbf v_{u,x}.
\end{align}

In our setting, computing Eq.~(\ref{eq:marginal_basic}) is intractable. Since $\mathbf v_{u,x}$ is a multi-dimensional value vector and each dimension corresponds to a 5-point Likert score, exact marginalization would require an exponential-time summation and is prohibitively expensive. This motivates us to employ variational inference to obtain a tractable lower bound for optimization.

% $o := (y_c \succ y_r)$ conditioned on $(\mathbf v_u, x)$:
% \begin{equation}
% \begin{aligned}
% \log p_{\theta}(o \mid \mathbf v_u, x)
% &=
% \log \int
% p_{\theta}(y_c \succ y_r \mid \mathbf v_{u,x}, x)
% \\
% &\quad\times
% p_{\theta}(\mathbf v_{u,x} \mid \mathbf v_u, x)
% \, d\mathbf v_{u,x}.
% \end{aligned}
% \label{eq:marginal_basic}
% \end{equation}

% \paragraph{Why the evidence is intractable}
% The integral in Eq.~(\ref{eq:marginal_basic}) marginalizes over the latent personalized values $\mathbf v_{u,x}$. In our setting, computing it exactly is intractable for several reasons. 
% First, $\mathbf v_{u,x}$ is a multi-dimensional value vector and each dimension corresponds to a 5-point Likert score, thus exact marginalization would require an exponential-time summation. Second, both the prior $p_{\theta}(\mathbf v_{u,x}\mid \mathbf v_u, x)$ and the preference likelihood $p_{\theta}(y_c \succ y_r \mid \mathbf v_{u,x}, x)$ are parameterized by neural/LLM-based modules, which generally yields a \emph{non-conjugate} model. Consequently, the posterior does not admit a closed-form normalization constant, and the evidence integral has no analytic solution.
% Finally, even numerically approximating Eq.~\eqref{eq:marginal_basic} via naive quadrature or exhaustive enumeration would be prohibitively expensive, since it would require repeatedly evaluating the likelihood
% over a vast latent space. This motivates using variational inference to obtain a tractable lower bound.

\paragraph{Variational posterior and ELBO.}
We introduce a variational posterior
$q_\phi(\mathbf v_{u,x}\mid \mathbf v_u, x)$ to approximate the true posterior $p_{\theta}(\mathbf v_{u,x}\mid \mathbf v_u, \rvx, o)$. Thus, we solve the optimization problem as:

\begin{equation}
\begin{aligned}
\hat{\phi}
&=
\operatorname*{arg\,min}_{\phi}\;
\mathrm{KL}\Bigl(
q_\phi(
    \mathbf v_{u,x}
    \mid \mathbf v_u,x
)
\\[-2pt]
&\qquad\qquad
\Big\Vert\;
p_\theta(
    \mathbf v_{u,x}
    \mid \mathbf v_u,x,o
)
\Bigr).
\end{aligned}
\label{app:variational_objective}
\end{equation}

In the following derivation, let
$\mathbf z := \mathbf v_{u,x}$ and use the abbreviations
\begin{align}
q_\phi(\mathbf z)
&:= q_\phi(\mathbf z \mid \mathbf v_u, x),
\nonumber\\
p_\theta(\mathbf z)
&:= p_\theta(\mathbf z \mid \mathbf v_u, x).
\nonumber
\end{align}
For brevity, we suppress the conditioning on
$(\mathbf v_u,x)$ below.

\begin{align}
&\mathrm{KL}\Bigl(
    q_\phi(\mathbf z)
    \,\Big\Vert\,
    p_\theta(\mathbf z\mid o)
\Bigr)
\notag\\
&=
\int q_\phi(\mathbf z)
\log
\frac{
    q_\phi(\mathbf z)
}{
    p_\theta(\mathbf z\mid o)
}
\,d\mathbf z
\notag\\
&=
\int q_\phi(\mathbf z)
\Bigl[
    \log q_\phi(\mathbf z)
    -
    \log p_\theta(\mathbf z\mid o)
\Bigr]
\,d\mathbf z
\notag\\
&=
\int q_\phi(\mathbf z)
\Bigl[
    \log q_\phi(\mathbf z)
    -
    \log p_\theta(\mathbf z,o)
\Bigr]
\,d\mathbf z
\notag\\
&\quad+
\log p_\theta(o)
\notag\\
&=
\log p_\theta(o)
\notag\\
&\quad-
\int q_\phi(\mathbf z)
\Bigl[
    \log p_\theta(\mathbf z,o)
    -
    \log q_\phi(\mathbf z)
\Bigr]
\,d\mathbf z .
\label{app:kl_objective}
\end{align}

We use $\mathcal{L}(q_{\phi})$ to represent the following term as:

\begin{align}
\mathcal{L}(q_{\phi})
&=
\int q_\phi
\bigl[
    \log p_{\theta}(\mathbf v_{u,x},o)
\notag\\
&\qquad
    -
    \log q_\phi(\mathbf v_{u,x})
\bigr]
\,d\mathbf v_{u,x}
\notag\\
&=
\mathbb{E}_{q_{\phi}}
\bigl[
    \log p_{\theta}(\mathbf v_{u,x},o)
\notag\\
&\qquad
    -
    \log q_\phi(\mathbf v_{u,x})
\bigr].
\label{app:elbo_term}
\end{align}

Based on Eq.(\ref{app:kl_objective}) and Eq.(\ref{app:elbo_term}), we then obtain
\begin{align}
    & \text{KL}(q_\phi(\mathbf v_{u,x}) || p_{\theta}(\mathbf v_{u,x}\mid o)) = \log p_{\theta}(o) - \mathcal{L}(q_{\phi}) \\
    & \log p_{\theta}(o) = \mathcal{L}(q_{\phi}) + \text{KL}(q_\phi(\mathbf v_{u,x}) || p_{\theta}(\mathbf v_{u,x}\mid o)).
\end{align}

Since $\log p_{\theta}(o)$ is fixed on the observed data, minimizing the KL divergence in Eq.(\ref{app:kl_objective}) is equivalent to maximizing $\mathcal{L}(q_{\phi})$, which is the evidence lower bound (ELBO) of $\log p_{\theta}(o)$.

\begin{align}\label{app:equal_loss}
\mathcal{L}(q_{\phi})
&=
\mathbb{E}_{q_{\phi}}
\bigl[
    \log p_{\theta}(\mathbf v_{u,x},o)
\notag\\
&\qquad
    -
    \log q_\phi(\mathbf v_{u,x})
\bigr]
\notag\\
&=
\mathbb{E}_{q_{\phi}}
\bigl[
    \log p_{\theta}(o|v_{u,x})
    +
    \log p_{\theta}(v_{u,x})
\notag\\
&\qquad
    -
    \log q_\phi(\mathbf v_{u,x})
\bigr]
\notag\\
&=
\mathbb{E}_{q_{\phi}}
\bigl[
    \log p_{\theta}(o|v_{u,x})
\bigr]
\notag\\
&\qquad
-
\mathbb{E}_{q_{\phi}}
\bigl[
    \log q_\phi(\mathbf v_{u,x})
    -
    \log p_{\theta}(v_{u,x})
\bigr]
\notag\\
&=
\mathbb{E}_{q_{\phi}}
\bigl[
    \log p_{\theta}(o|v_{u,x})
\bigr]
\notag\\
&\qquad
-
\text{KL}\bigl(
    q_\phi(\mathbf v_{u,x})
    ||
    p_{\theta}(v_{u,x})
\bigr).
\end{align}
Converting to the training loss, we need to minimize the loss $\mathcal{L}$ as follows:
\begin{align}
\mathcal{L}
&=
-
\mathbb{E}_{q_{\phi}}
\bigl[
    \log p_{\theta}(o|v_{u,x})
\bigr]
\notag\\
&\qquad
+
\text{KL}\bigl(
    q_\phi(\mathbf v_{u,x})
    ||
    p_{\theta}(v_{u,x})
\bigr),
\end{align}
which is Eq.(\ref{eq:elbo_loss}) in the main paper.

\subsection{Concrete Formulation}
\label{app:formula}

\paragraph{Setup.}
For a user $u$ and a question $x$, let $\mathbf v_u$ be the user-level value profile (Sec.~\ref{subsec:task_definition}), $\mathbf v_x$ be the scenario value-salience variable estimated from $(x, y_{\text{human}})$ (Sec.~\ref{subsec:scenario_value_salience}),
and $\mathbf v_{u,x}$ be the question-specific personalized values. Given a preference triple $(x,y_c,y_r)$, we extract the value vector reflected by both the preferred and dispreferred responses as $v_{y_c}, v_{y_r}$. 

% we extract a \emph{single-response} value representation
% \begin{equation}
% \mathbf v_a := f_{\mathrm{eval}}(y_c),
% \end{equation}
% where $f_{\mathrm{eval}}(\cdot)$ is the value extractor used in our pipeline.
% The dispreferred response $y_r$ is used only to provide negatives for contrastive learning
% (Sec.~3.4), rather than defining a likelihood term. Here, $\mathbf v_a$ serves as the \emph{observed training signal},
% constructed by applying the value extractor to the preferred response $y_c$. $\mathbf v_a$ provides a supervision target during training, enabling the learned
% $\mathbf v_{u,x}$ to align with the preference relation $y_c \succ y_r$
% when evaluated on held-out examples.

\paragraph{Generative model (hierarchical prior).}
We use the estimated value salience distribution
$p_\omega(\mathbf v_x \mid x, y_{\text{human}})$ and a user-conditioned prior
$p_\theta(\mathbf v_{u,x} \mid \mathbf v_u, \mathbf v_x)$.
We further introduce a (single-point) observation model for the extracted value embedding:
\begin{equation}
p_\theta(\mathbf v_{y_c} \mid \mathbf v_{u,x}).
\end{equation}
The resulting marginal likelihood is
\begin{equation}
\begin{aligned}
\log p(\mathbf v_{y_c} \mid \mathbf v_u, x)
&= \log \sum_{\mathbf v_x} \sum_{\mathbf v_{u,x}}
p_\theta(\mathbf v_{y_c} \mid \mathbf v_{u,x}) \\
&\qquad \cdot p_\theta(\mathbf v_{u,x} \mid \mathbf v_u, \mathbf v_x) \\
&\qquad \cdot p_\omega(\mathbf v_x \mid x, y_{\text{human}}).
\end{aligned}
\label{eq:marginal_with_vx}
\end{equation}

\subsection{Prompt Template}
\label{prompt_template}
We present below the prompts used in this paper, including the data preprocessing prompt, the method prompt, and the evaluation prompt.

\subsubsection{Preprocessing Prompt}
\label{Preprocessing_prompt}
This stage involves four preprocessing prompts:
(1) extracting the user value profile in Fig.~\ref{survey_prompt},
(2) determining whether a given question involves value-related considerations in Fig.~\ref{question_prompt},
(3) identifying which values are implicated by the question in Fig.~\ref{question_relevant_prompt}, and
(4) extracting value-related representations from candidate answers in Fig.~\ref{eval_prompt}.

\begin{figure*}[htbp]
\centering
\small

\begin{tcolorbox}[
% verbatim, % <--- Remove this line
center,
title=Prompt: Schwartz Basic Human Values Evaluation on Personal Profile.,
text width=\linewidth,
boxrule = 1.5pt,
boxsep=2mm,
% fontbody=\ttfamily % Optional: to make the text look like verbatim
]

\textbf{Background:} \\
You are an AI assistant tasked with evaluating how strongly Schwartz's basic human values are reflected in a given personal description. The input text is a self-description written by a person, and your goal is to infer how strongly each of the ten Schwartz values appears in what the person expresses, implies, or emphasizes.

\textbf{Schwartz’s Ten Basic Values:}
Self-direction — independent thought and action; choosing, creating, exploring

Stimulation — excitement, novelty, and challenge in life

Hedonism — pleasure or sensuous gratification for oneself

Achievement — personal success through demonstrating competence according to social standards

Power — social status and prestige, control or dominance over people and resources

Security — safety, harmony, and stability of society, relationships, and of self

Conformity — restraint of actions, inclinations, and impulses that may upset or harm others or violate social expectations

Tradition — respect, commitment, and acceptance of cultural or religious customs and ideas

Benevolence — preserving and enhancing the welfare of people with whom one is in frequent personal contact

Universalism — understanding, appreciation, tolerance, and protection for the welfare of all people and for nature

\textbf{Scoring Scale (1--5):} \\
1 (Contradicted) — The text denies, criticizes, or opposes this value. \\
2 (Absent) — The value is not mentioned, implied, or relevant in any way. \\
3 (Mentioned but Not Important) — The value is referenced or implied, but it is not meaningful to the person's identity or motivations. \\
4 (Present but Not Central) — The value seems to matter to the person and influences their perspective, but it is not central. \\
5 (Most Important) — The value is central to the person's identity or worldview. Removing it would significantly change their self-description. \\

\vspace{0.5em}
\textbf{Task:} \\
Evaluate the following personal self-description and rate how strongly each of the ten Schwartz values is reflected in it.
\\

\textbf{Input Text:} \\
\texttt{"\{personal\_text\}}

\vspace{0.5em}
\textbf{Output Format (STRICT):} \\
Rate all ten values using only integers 1--5. Output ONLY in the following format:

\begin{verbatim}
Self-direction: [score]
Stimulation: [score]
Hedonism: [score]
Achievement: [score]
Power: [score]
Security: [score]
Conformity: [score]
Tradition: [score]
Benevolence: [score]
Universalism: [score]
\end{verbatim}

\end{tcolorbox}
\caption{Schwartz basic human values evaluation on personal profile prompt.}
\label{survey_prompt}
\end{figure*}

\begin{figure*}[htbp]
\centering
\small

\begin{tcolorbox}[
% verbatim, % <--- Remove this line
center,
title=Prompt: Schwartz Value Candidate Identification.,
text width=\linewidth,
boxrule = 1.5pt,
boxsep=2mm,
% fontbody=\ttfamily % Optional: to make the text look like verbatim
]

\textbf{Background:} \\
You are an AI assistant that identifies which Schwartz basic human values are relevant to a given text. Your task is to determine which values are implied, expressed, or thematically connected to the provided {text\_type}, and output them in a structured JSON format.

\textbf{Schwartz’s Ten Basic Values:}
Self-direction — independent thought and action; choosing, creating, exploring

Stimulation — excitement, novelty, and challenge in life

Hedonism — pleasure or sensuous gratification for oneself

Achievement — personal success through demonstrating competence according to social standards

Power — social status and prestige, control or dominance over people and resources

Security — safety, harmony, and stability of society, relationships, and of self

Conformity — restraint of actions, inclinations, and impulses that may upset or harm others or violate social expectations

Tradition — respect, commitment, and acceptance of cultural or religious customs and ideas

Benevolence — preserving and enhancing the welfare of people with whom one is in frequent personal contact

Universalism — understanding, appreciation, tolerance, and protection for the welfare of all people and for nature

\vspace{0.5em}
Identify up to 5 relevant Schwartz values for the {text\_type}. \\
Use only these keys:

\begin{verbatim}
['self_direction', 'stimulation', 'hedonism', 'achievement', 'power',
 'security', 'conformity', 'tradition', 'benevolence', 'universalism']
\end{verbatim}

For each selected value, provide a relevance score between 0 and 1.

\vspace{0.5em}
\textbf{Output Format (STRICT JSON ONLY):} \\
Return only valid JSON in exactly the following structure:

\begin{verbatim}
{"candidates":[
  {"value":"<one_of_keys>","relevance":0.0-1.0}
]}
\end{verbatim}

\end{tcolorbox}
\caption{Prompt for Schwartz value candidate identification of question.}
\label{question_prompt}
\end{figure*}

\begin{figure*}[htbp]
\centering
\small

\begin{tcolorbox}[
% verbatim, % <--- Remove this line
center,
title=Prompt: Prompt: Value-Related Question Identification.,
text width=\linewidth,
boxrule = 1.5pt,
boxsep=2mm,
% fontbody=\ttfamily % Optional: to make the text look like verbatim
]

\textbf{Background:} \\
You are an expert in Schwartz's Theory of Basic Human Values. Your task is to determine whether a given question or statement is related to human values. A question is value-related if it reflects moral priorities, ethical considerations, social judgments, or underlying motivational principles connected to Schwartz’s ten basic values.

\textbf{Schwartz’s Ten Basic Human Values:}

\texttt{"\{value definition\}}

\textbf{A question is considered value-related if:}
\begin{itemize}
\item It asks about moral or ethical choices.
\item It involves trade-offs between different priorities (e.g., personal gain vs. helping others).
\item It touches on topics such as fairness, freedom, security, achievement, pleasure, tradition, caring for others, or environmental protection.
\item It asks opinions about social, political, or personal matters that reflect underlying values.
\item It involves decision-making influenced by value priorities.
\end{itemize}

\textbf{A question is NOT value-related if:}
\begin{itemize}
\item It is purely factual.
\item It is a technical or procedural question.
\item It asks for objective information without value judgment.
\end{itemize}

\vspace{0.5em}
\textbf{Input Question:} \\
\texttt{"\{question\}}

\vspace{0.5em}
\textbf{Output Requirement:} \\
Respond with ONLY:

\texttt{True} — if the question is value-related.

\texttt{False} — if the question is not value-related.

\end{tcolorbox}
\caption{Prompt for value-related question identification.}
\label{question_relevant_prompt}
\end{figure*}

\begin{figure*}[htbp]
\centering
\small

\begin{tcolorbox}[
% verbatim, % <--- Remove this line
center,
title=Prompt: Answer Values Evaluation.,
text width=\linewidth,
boxrule = 1.5pt,
boxsep=2mm,
% fontbody=\ttfamily % Optional: to make the text look like verbatim
]

\textbf{Background:} \\
You are an AI assistant tasked with evaluating how strongly Schwartz's basic human values are reflected in a given personal description. The input text is a self-description written by a person, and your goal is to infer how strongly each of the ten Schwartz values appears in what the person expresses, implies, or emphasizes.

\textbf{Schwartz’s Ten Basic Values:}

\texttt{"\{value definition\}}

\vspace{0.5em}
\textbf{Scoring Scale (1--5):} \\
1 (Contradicted) — The text denies, criticizes, or opposes this value. \\
2 (Absent) — The value is not mentioned, implied, or relevant in any way. \\
3 (Mentioned but Not Important) — The value is referenced or implied, but it is not meaningful to the person's identity or motivations. \\
4 (Present but Not Central) — The value seems to matter to the person and influences their perspective, but it is not central. \\
5 (Most Important) — The value is central to the person's identity or worldview. Removing it would significantly change their self-description. \\

\vspace{0.5em}
\textbf{Task:} \\
Evaluate the following personal self-description and rate how strongly each of the ten Schwartz values is reflected in it.

\textbf{Input Text:} \\
\texttt{"\{text\}}

\vspace{0.5em}
\textbf{Output Format (STRICT):} \\
Rate all ten values using only integers 1--5. Output ONLY in the following format:

\begin{verbatim}
Self-direction: [score]
Stimulation: [score]
Hedonism: [score]
Achievement: [score]
Power: [score]
Security: [score]
Conformity: [score]
Tradition: [score]
Benevolence: [score]
Universalism: [score]
\end{verbatim}

\end{tcolorbox}
\caption{Prompt for answer value evaluation.}
\label{eval_prompt}
\end{figure*}

\subsubsection{Methods Prompt}
\label{Methods_prompt}

\paragraph{View-Specific Residual Inference Prompts}
To instantiate the residual distributions $q(\rvz_{\rvu})$ and $q(\rvz_{\rvx})$ in Eq.~(6) and Eq.~(8), 
we implicitly estimate the residual corrections conditioned on the triplet $(x, \rvv_x, \rvv_{u_i})$. 
Rather than explicitly sampling $\rvz$, we adopt a prompt-based inference strategy in Fig.~\ref{question_value_infer} to enable the model to 
capture how contextual semantics modulate value deviations.

Specifically, we first construct a structured prompt that instructs the language model to analyze the input 
question $x$ from the perspective of Schwartz’s ten basic human values. 
This prompt is designed to emphasize value salience under the given scenario, allowing the model to 
encode context-sensitive value implications. 
The resulting hidden representation is extracted as a contextual embedding $\rve_x$, which serves as 
a semantic anchor for residual estimation.

Conditioned on $\rve_x$ and the corresponding prior (either $\rvv_x$ for the scenario-anchored view or 
$\rvv_{u_i}$ for the person-anchored view), a lightweight projection network produces a logit-level 
correction that implicitly instantiates the residual variable $\rvz$. 
This correction is defined in the same value space as the prior distribution and is subsequently used 
to refine the posterior value inference. 
In this way, the effect of $x$ is captured through semantic prompting rather than 
explicit stochastic modeling.

\paragraph{Gating Prompt for View Composition}
To compute the adaptive fusion weight $\alpha$ in Eq.~(10), we employ a prompt-based gating mechanism. 
Given the input question $x$ and the predictions from both views, we construct a concise comparison prompt in Figure~\ref{gate_prompt}
that presents the value-wise predictions and their associated confidence signals. 
The prompt instructs the language model to decide which view provides a more reliable assessment under the 
current context.

The gating model outputs logits corresponding to two discrete choices (scenario-anchored vs.\ person-anchored). 
We extract the logits associated with tokens \texttt{0} and \texttt{1}, and normalize them to obtain the 
scalar gating coefficient $\alpha \in (0,1)$. 
This value is then used to interpolate between the two inferred posteriors, enabling context-aware 
view composition without introducing additional heuristic rules.

\begin{figure*}[htbp]
\centering
\small

\begin{tcolorbox}[
center,
title=Prompt: Question Information Extract for Latent Variable.,
text width=\linewidth,
boxrule = 1.5pt,
boxsep=2mm,
]

\textbf{Instruction:} \
Analyze the following question from the perspective of Schwartz's Theory of Basic Human Values.

\textbf{Value Dimensions:} \
Consider the 10 value dimensions: Self-Direction, Stimulation, Hedonism, Achievement, Power,
Security, Conformity, Tradition, Benevolence, and Universalism.

\vspace{0.5em}
\textbf{Question:} \texttt{"\{question\}"}

\end{tcolorbox}

\caption{Prompt for Schwartz value-dimension analysis of question.}
\label{question_value_infer}
\end{figure*}

\begin{figure*}[htbp]
\centering
\small

\begin{tcolorbox}[
% verbatim, % <--- Remove this line
center,
title=Prompt: Gate Function.,
text width=\linewidth,
boxrule = 1.5pt,
boxsep=2mm,
% fontbody=\ttfamily % Optional: to make the text look like verbatim
]

\textbf{Background:} \
You are a value alignment meta-evaluator. Your goal is to compare two prediction streams and decide which one gives a more rational and reliable value-based assessment for a given question. The two streams are: \

\textbf{Stream 0 (Universal):} Reflects common societal values and general logic. \

\textbf{Stream 1 (Personal):} Reflects specific user traits and historical profile. \

You will be given a comparison matrix evaluating both streams along key value dimensions.

\vspace{0.5em}
\textbf{Task:} \
Choose which prediction stream (Universal vs. Personal) provides a \emph{more rational} assessment for the primary value dimensions of the question. You must output either \texttt{0} (for Universal) or \texttt{1} (for Personal). \

\textbf{Reliability Tags:} \

\textbf{[Reliable]}: High confidence in the stream's prediction quality and stability. \

\textbf{[Uncertain]}: Low confidence, noisy or unstable prediction. \

\vspace{0.5em}
\textbf{Input:} \

\textbf{Question:} \texttt{"{question}"}

\textbf{Comparison Matrix (Scale 1--5):}
\begin{verbatim}
Dimension    Stream 0 (Universal)    
Stream 1 (Personal)
{comparison_table}
\end{verbatim}

\textbf{Scale Meaning (1--5):} \
1 (Contradicted): The value is strongly violated or opposed. \
2 (Absent): The value is missing or ignored. \
3 (Mentioned but Not Important): The value appears but is not emphasized. \
4 (Present but Not Central): The value is clearly present but not dominant. \
5 (Most Important): The value is central and primary in reasoning. \

\vspace{0.5em}
\textbf{Analytical Guidance:} \

Identify which value dimensions are \emph{primary} for evaluating this question (e.g., fairness, harm, autonomy, etc.). \

For each primary dimension, compare the scores of Stream 0 vs. Stream 1. \

Consider: \
\quad- Does Universal reasoning better match broad, widely accepted norms? \
\quad- Does Personal reasoning risk overfitting to idiosyncratic user traits? \
\quad- Are there dimensions where one stream is clearly unreliable (very low scores)? \

Integrate across all primary dimensions to decide which stream gives the overall more rational and trustworthy value assessment. \

\vspace{0.5em}
\textbf{Output Format (STRICT):} \
You must output \emph{only} a single number with no explanation: \
\begin{itemize}
\item Output \texttt{0} if Stream 0 (Universal) provides the more rational assessment. \
\item Output \texttt{1} if Stream 1 (Personal) provides the more rational assessment. \
\end{itemize}

\textbf{Final Question:} \
Which stream provides a more rational assessment for the primary value dimensions of this question? \

\textbf{Answer:}
\begin{verbatim}
0
\end{verbatim}
(or)
\begin{verbatim}
1
\end{verbatim}
\end{tcolorbox}
\caption{Prompt for gate function.}
\label{gate_prompt}
\end{figure*}

\subsubsection{Evaluation Prompt}

The prompt in Fig.~\ref{eval_prompt} is also used to evaluate responses generated by some baseline models,
as described in Appendix~\ref{app:baselines}.

\section{Additional Experimental Details}
\label{app:exp_setup}

\subsection{Datasets and Preprocessing}
\label{app:datasets}

\paragraph{PRISM.}
We use the PRISM dataset \citep{kirk2024prism} to evaluate value-conditioned personalization and
value alignment in open-ended user--model interactions.
Each instance is organized turn-by-turn, containing (i) a user query and its conversation context,
(ii) a set of candidate responses provided by the dataset (including the chosen response and up to three unselected responses),
and (iii) user-side survey metadata (e.g., \texttt{self\_description}) when available.
In our formulation, $x$ denotes the full user query context (the user message concatenated with the \texttt{opening\_prompt} for the first turn),
and $v_u$ denotes the user value profile derived from survey text (details below).

\paragraph{GOOD (P4G-derived).}
Our GOOD dataset is constructed from the PersuasionForGood (P4G) corpus \citep{wang-etal-2019-persuasion},
a persuasion dialogue benchmark collected via Amazon Mechanical Turk.
Each dialogue involves a \emph{Persuader} and a \emph{Persuadee}, where the Persuader aims to persuade the Persuadee to donate to a charity (e.g., Save the Children).
The original corpus contains 1,017 complete dialogues and rich participant metadata including Schwartz values (10 dimensions), Big-Five personality traits,
and Moral Foundations.

\subsubsection{Value Extraction and Alignment Scoring}
\label{appendix:data_preprocessing}

To quantify the alignment between user values and model responses on PRISM \citep{kirk2024prism}, we develop a pipeline that produces value scores for (i) the user profile $v_u$ and (ii) model responses $v_{y_c}$.

\paragraph{Selection and relevance filtering.}
We focus on the \texttt{values guided} subset, where models are explicitly prompted to engage with value-laden topics. To ensure value salience and data quality, we first retain only conversations with high value relevance, specifically those with \texttt{choice\_attributes.values} scores of at least 80. We then perform turn-level processing, handling each conversation sequentially. Finally, we keep only users who provide survey text in \texttt{self\_description}, which enables extraction of the user value profile $v_u$.

\paragraph{LLM-based value extraction.}
We use an LLM as an evaluator to quantify the 10 basic human values
(Self-direction, Stimulation, Hedonism, Achievement, Power, Security, Conformity, Tradition, Benevolence, Universalism).
For each turn:
\begin{enumerate}
    \item \textbf{Relevance classification:} classify whether the query is value-related; discard non-relevant turns.
    \item \textbf{Candidate dimension identification:} select up to 5 value dimensions most pertinent to the query.
    \item \textbf{Value scoring (user vs. model):}
    \begin{itemize}
        \item \textbf{User profile ($v_u$):} score each dimension (1--5) from \texttt{self\_description}.
        \item \textbf{Chosen answer ($v_{y_c}$):} score the model's chosen response across all 10 dimensions.
        \item \textbf{Unselected answers:} score up to 3 unselected candidate responses to serve as comparison baselines.
    \end{itemize}
\end{enumerate}

\subsubsection{Evaluation Metrics}
We compute three metrics. Since the GOOD dataset does not contain negative samples, accuracy is only reported for PRISM.
 
\begin{itemize}
    \item \textbf{MAE (value intensity mismatch):} for each candidate dimension $d$,
    $\mathrm{MAE}_d = |\hat{v}_{u_i,x}^d -v_{y_c}^d|$.
    
    \item \textbf{Spearman’s rank correlation coefficient (Correlation):} we use spearman correlation metric:
    \begin{equation}
\rho = 1 - \frac{6 \sum_{i=1}^{n} d_i^2}{n(n^2 - 1)},
\end{equation}
    
    \item \textbf{Accuracy}: Given one positive candidate and one or more negative candidates,
this metric measures whether the predicted alignment is closer to the positive sample
than to the negative samples. Formally, let $\hat{v}_{u_i,x}$ denote the predicted alignment
vector, $v_{y_c}$ the alignment vector of the positive sample, and
$v_{y_r}$ that of a negative sample. The prediction is considered correct if
\begin{equation}
\left\lVert \hat{v}_{u_i,x} - v_{y_c} \right\rVert < \left\lVert \hat{v}_{u_i,x} - v_{y_r} \right\rVert .
\end{equation}

\end{itemize}

\subsection{Baselines}
\label{app:baselines}
We compare our method with a diverse set of baselines covering both
\emph{non-personalized} and \emph{personalized} value alignment paradigms.
All personalized baselines incorporate user value information at inference time,
while non-personalized baselines do not utilize any user-specific value signals.

\paragraph{Non-personalized Baselines.}
These methods generate responses without explicitly incorporating user value profiles.

\textbf{DirectAnswer} directly generates responses by conditioning the language model on the user query and conversational context $x$, without considering any user-specific value information. This baseline represents general value alignment learned during pretraining.

\paragraph{Personalized Baselines.}
These methods leverage user value information to produce personalized, value-aligned responses.

\textbf{Value Prompt} explicitly injects the user value profile $v_u$ into the system prompt, steering generation toward user-aligned value expression.

\textbf{MetaAligner} performs meta-level alignment by rewriting or steering generated responses toward preferred values inferred from the user value profile and contextual information.

\textbf{COUPLE} employs explicit preference modeling and counterfactual reasoning to encourage consistency between generated responses and the user value profile.

\textbf{PAD} and \textbf{MOD} apply preference-aware decoding and modulation strategies to bias token selection toward user-aligned values during inference.

\textbf{ValuesRAG} retrieves historical user cases with similar scenarios and value profiles as examples to support personalized decision-making.

\textbf{PersonValue} is implemented by us and compares the static user value profile $v_u$ with values extracted from candidate responses, serving as a static personalization baseline without scenario-aware inference.

As many baseline methods produce answers as their final outputs, we evaluate value alignment by extracting value representations from the generated responses. The specific prompts used for value extraction are detailed in the Appendix.

\subsection{Implementation Details}

\paragraph{Backbone and Training.}
We adopt the Qwen3-8B model as the backbone and fine-tune it using Q-LoRA for parameter-efficient training. The model is optimized with Adam using a learning rate of $5 \times 10^{-3}$ and a batch size of 16. Training is conducted for two epochs with a fixed random seed of 42 to ensure reproducibility. During the second epoch, all parameters except the gating module are frozen, allowing the training process to focus on optimizing the $\alpha$ parameters.

\paragraph{Hyperparameters.}
We introduce a hyperparameter $\lambda_1$ to weight the KL-divergence term in our objective.
$\lambda_1$ is searched within the range $[0,1]$, and the final value is selected based on validation performance.
Meanwhile, the gate coefficient $\alpha$ is initialized to 0.5 to provide a balanced prior contribution at the beginning of training.

\subsection{Human Evaluation}
\label{app:human_eval}

In this section, we present the details of the human evaluation. We conducted two rounds of human evaluation: the first was to verify the effectiveness of the evaluator used in $f_{eval}$, and the second was to validate the performance of our proposed method.

\begin{figure*}[t]
    \centering
    \includegraphics[width=\linewidth]{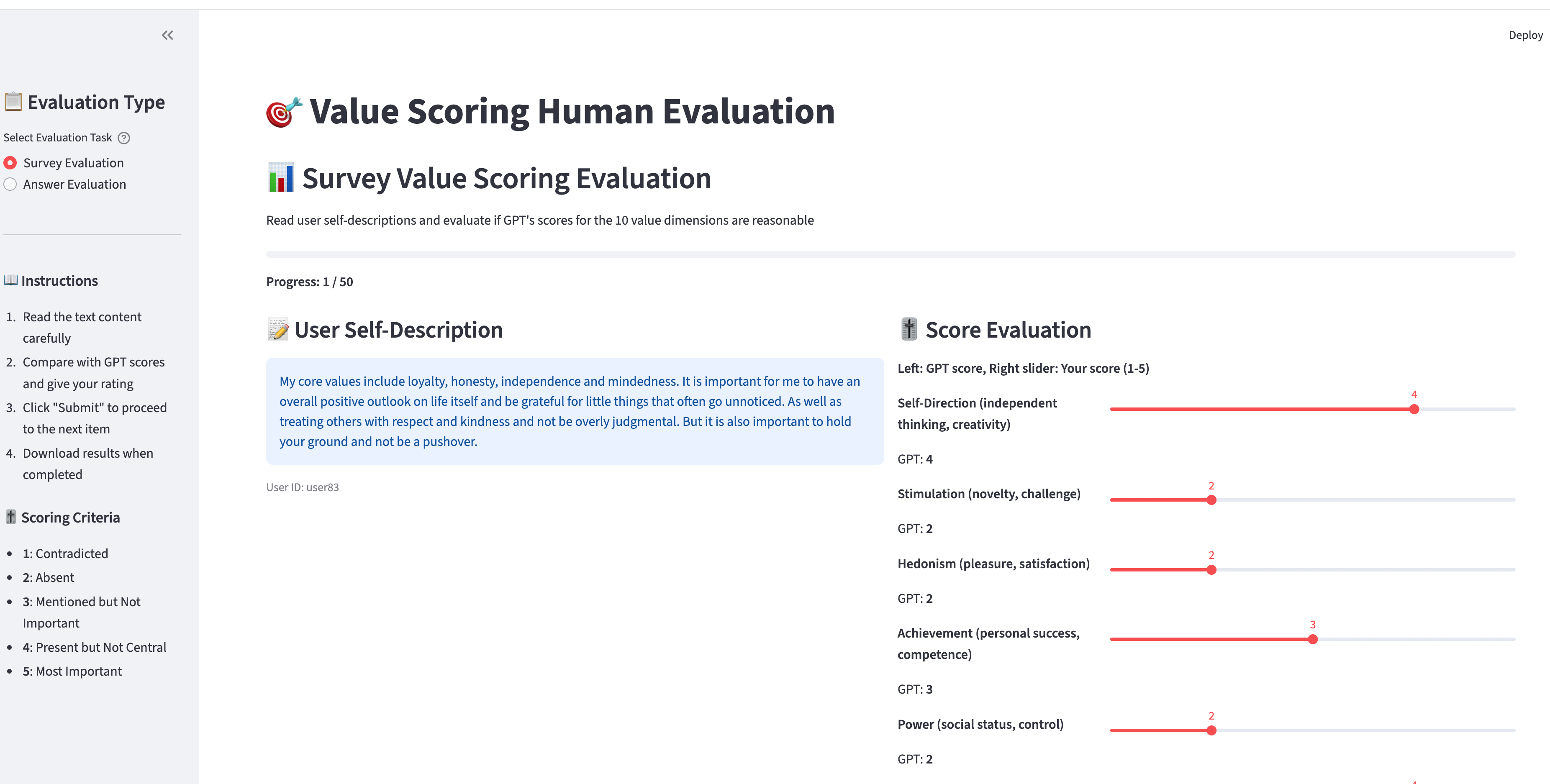}
    \caption{Human annotation interfaces used in the first human evaluation. Annotators independently label values based on a given user profile derived from a survey or a given question--answer pair.}
    \label{fig:human_eval_1}
\end{figure*}

\paragraph{Evaluator Validation.}
We conduct a human evaluation to examine whether the evaluator employed in $f_{\text{eval}}$ is consistent with human judgments. To this end, we design two independent annotation settings. In the first setting, annotators are provided with a user profile derived from a survey and are asked to label the underlying values reflected in the profile. In the second setting, annotators are given a question--answer pair and are asked to annotate the values expressed in the answer. The evaluator-generated scores are then compared with these human annotations. This annotation interface is shown in Fig.~\ref{fig:human_eval_1}.

As summarized in Tab.~\ref{tab:evaluator_validation}, the evaluator demonstrates a high level of agreement with human judgments across both settings, achieving low mean absolute error (MAE), high Spearman correlation, and strong agreement rates. These results are consistent with prior findings reported in COUPLE~ \citep{guo2026counterfactual}, indicating that large language model–based evaluators can reliably extract human values and serve as an effective proxy for human evaluation.

\begin{table}[htbp]
\centering
\small
\resizebox{\linewidth}{!}{%
\begin{tabular}{lcccc}
\toprule
\textbf{Setting} & \textbf{MAE} & \textbf{Corr $r$} & \textbf{Acc. ($\leq$1)} & \textbf{Exact Match} \\
\midrule
Survey-level     & 0.056 & 0.986 & 100.0\% & 94.4\% \\
Answer-level     & 0.120 & 0.936 & 98.4\%  & 89.6\% \\
\bottomrule
\end{tabular}
}
\caption{Agreement between the automatic evaluator and human annotations under survey-level and answer-level evaluation settings.}
\label{tab:evaluator_validation}
\end{table}

\paragraph{Model Performance Evaluation.}
The second human evaluation is conducted to assess the effectiveness of our proposed method in predicting context-specific personalized values. We sample 50 instances corresponding to 50 distinct survey-derived user profiles, with each profile appearing in exactly one instance. The complete evaluation set is independently assessed by two human evaluators, both of whom review all 50 samples under identical instructions. For each sample, annotators are presented with the survey-derived user profile, the question, the target answer, and the anonymized value-dimension assessments produced by BaCVA and ValuesRAG. Following the same five-level rubric, they determine which assessment more accurately reflects the values expressed in the target answer. When both evaluators prefer the same method, the sample is counted as a win or loss for BaCVA; when they disagree, it is counted as a tie. The annotation examples are shown in Tab.~\ref{tab:results_analysis_example}.

\begin{table*}[t]
\centering
\small
\setlength{\tabcolsep}{3pt}
\begin{tabular}{p{2cm} p{6cm} c c c c c}
\toprule
\textbf{Question} & \textbf{Answer} & \textbf{\#Dims} & \textbf{Security} & \textbf{Conformity} & \textbf{Benevolence} & \textbf{Chosen} \\
\midrule
What to do if mum remains dead-set on her opinions? 
& If your mum remains dead-set on her opinions, it's important to approach the situation with empathy and understanding. Here are a few suggestions: acknowledge her concerns, offer compromise, seek support from a nutritionist, and encourage open communication.
& 3
& 4.0 / 4.0
& 4.0 / 4.0
& 5.0 / 4.0
& A or B \\
\bottomrule
\end{tabular}
\caption{Example results of blind value dimension analysis. Scores are reported as \textit{Model A / Model B} for each value dimension, and \textit{Chosen} indicates the model preferred by human annotators.}
\label{tab:results_analysis_example}
\end{table*}

\paragraph{Annotator Details.}
All annotation procedures were approved by the relevant Institutional Review Board (IRB) and conducted in accordance with established ethical research guidelines. Annotators were informed about the task scope and voluntarily consented to participate. All annotators were compensated at a rate of at least \$7 per hour, which substantially exceeds the prevailing local hourly wage levels in the countries where they were employed. The annotation tasks did not involve any personal, private, or sensitive information. To ensure independence and avoid potential bias, the two human evaluation tasks were conducted with two separate groups of annotators, with no overlap between the annotators involved in each task. All annotators had a background in psychology, received training on Schwartz’s theory of basic human values, and were proficient in English reading comprehension.

% \paragraph{Human Evaluation Results.}

% As shown in Fig.~\ref{fig:human}, BaCVA is preferred over ValuesRAG in human evaluation, achieving a higher win rate and fewer losses. 
% This indicates that BaCVA does not merely improve automatic value-alignment metrics, but also produces responses that human annotators judge as more consistent with user preferences and contextual value needs.

\subsection{Licenses for Existing Assets}
\label{sec:licenses_existing_assets}

We use two existing datasets in our work: PRISM and GOOD. More information about the source datasets and their licenses is provided below.

\begin{itemize}
    \item \textbf{PRISM}~\citep{kirk2024prism}:
    Human-written texts (including prompts) in the PRISM dataset are licensed under the
    Creative Commons Attribution 4.0 International License (CC-BY-4.0). Model responses
    are licensed under the Creative Commons Attribution--NonCommercial 4.0 International
    License (CC-BY-NC-4.0). Use of model responses must also comply with the original
    licenses of the corresponding model providers.

    \item \textbf{GOOD (P4G-derived)}~\citep{wang-etal-2019-persuasion}:
    The GOOD dataset used in this work is derived from the PersuasionForGood (P4G) corpus,
    which is released under the Apache License 2.0. All derived data retain the original
    license of the P4G dataset, and appropriate attribution to the original source is
    required.
\end{itemize}

All source datasets used in our benchmark retain their original licenses, as specified
by their respective creators.

\section{Additional Experiments}
\label{app:addition_exp}
\subsection{Hyperparameter Sensitivity Analysis}
\label{app:hy_exp}
We analyze the sensitivity of BaCVA to key design choices by conducting hyperparameter studies.
% Specifically, we examine how the choice of backbone model for generating the universal response affects alignment performance, and whether BaCVA remains robust across different value systems used for evaluation.

\paragraph{Effect of KL Weight $\lambda_1$.}
We further study the effect of the KL-divergence weight $\lambda_1$, which controls the regularization strength between the variational posterior and the personalized prior. Tab.~\ref{tab:lambda_sensitivity} reports the sensitivity results on PRISM and GOOD. Overall, a moderate KL weight improves performance over removing the KL term, suggesting that the personalized prior provides useful regularization. However, overly large values of $\lambda_1$ lead to clear degradation, as the posterior is over-regularized toward the prior and becomes less adaptive to the current scenario. These results show that BaCVA is robust within a moderate range of $\lambda_1$, while very strong KL regularization is harmful.

\begin{table}[t]
\centering
\small
\caption{Sensitivity analysis of the KL-divergence weight $\lambda_1$ on PRISM and GOOD. Lower MAE is better, while higher correlation and accuracy are better.}
\label{tab:lambda_sensitivity}
\resizebox{\linewidth}{!}{%
\begin{tabular}{lccc|lcc}
\toprule
\multicolumn{4}{c|}{\textbf{PRISM}} & \multicolumn{3}{c}{\textbf{GOOD}} \\
\midrule
$\lambda_1$ & MAE $\downarrow$ & Corr. $\uparrow$ & Accuracy $\uparrow$
& $\lambda_1$ & MAE $\downarrow$ & Corr. $\uparrow$ \\
\midrule
0      & 0.7848 & 0.672 & 80.26\% & 0      & 1.3064 & 0.718 \\
1e-2   & 0.7572 & 0.678 & 80.59\% & 1e-2   & 1.2537 & 0.732 \\
3e-2   & 0.7548 & 0.696 & 81.68\% & 3e-2   & 1.1723 & 0.749 \\
1e-1   & 0.7344 & 0.703 & 81.39\% & 1e-1   & 1.1230 & 0.765 \\
3e-1   & 0.7809 & 0.601 & 77.44\% & 3e-1   & 1.1429 & 0.756 \\
1      & 0.9405 & 0.523 & 76.30\% & 1      & 1.3885 & 0.683 \\
\bottomrule
\end{tabular}%
}
\end{table}

\subsection{Comparison with Training-Time Alignment Methods}
\label{app:training_time_baselines}

To further compare BaCVA with training-time alignment methods, we include RLHF and MORLHF as additional baselines.
All methods use the same backbone, data splits, preference samples, and input information for a controlled comparison.
Standard RLHF learns a unified preference objective from the pooled preference data, whereas MORLHF performs multi-objective optimization over value-related preference signals.
The detailed training configurations follow the same experimental budget used for the main baselines.

As shown in Tab.~\ref{tab:training_time_baselines}, BaCVA achieves the best performance across all three metrics.
Compared with the strongest baseline for each metric, BaCVA reduces MAE by 62.3\% relative to MORLHF, while improving correlation by 0.191 and accuracy by 12.72 percentage points over MetaAligner.

\begin{table}[t]
\centering
\small
\setlength{\tabcolsep}{5pt}
\renewcommand{\arraystretch}{1.1}
\begin{tabular}{lccc}
\toprule
\textbf{Method}
& \textbf{MAE} $\downarrow$
& \textbf{Corr.} $\uparrow$
& \textbf{Accuracy} $\uparrow$ \\
\midrule
RLHF          & 2.969 & 0.312 & 48.92\% \\
MORLHF        & 1.933 & 0.446 & 53.41\% \\
MetaAligner   & 1.955 & 0.509 & 69.00\% \\
BaCVA         & \textbf{0.728} & \textbf{0.700} & \textbf{81.72\%} \\
\bottomrule
\end{tabular}
\caption{Comparison with training-time alignment methods on PRISM.}
\label{tab:training_time_baselines}
\end{table}

\begin{table}[t]
\centering
\small
\setlength{\tabcolsep}{6pt}
\renewcommand{\arraystretch}{1.08}
\caption{Prompt sensitivity analysis of the universal response estimator. 
Agreement is computed with the default Origin prompt as reference.}
\label{tab:prompt_sensitivity}
\begin{tabular}{lcc}
\toprule
Estimator Prompt & Agreement Corr. & Final MAE $\downarrow$ \\
\midrule
Origin & 1.000 & 0.728 \\
More Natural & 0.806 & \textbf{0.715} \\
Structural & 0.738 & 0.743 \\
\bottomrule
\end{tabular}
\end{table}

\paragraph{Effect of Prompt Designs.}
We further examine whether the estimation of the scenario-based prior is sensitive to prompt formulation.
Specifically, we compare three prompting strategies for generating the universal response: 
\emph{Origin}, which directly answers the input question; 
\emph{More Natural}, which encourages fluent and conversational responses; 
and \emph{Structural}, which produces structured, point-by-point narratives.
As shown in Tab.~\ref{tab:prompt_sensitivity}, although different prompts lead to moderate variations in inter-prompt agreement, the final downstream MAE remains stable across prompt designs.
Notably, the more natural prompt even slightly improves MAE from 0.728 to 0.715, while the structural prompt only introduces a minor degradation to 0.743.
These results indicate that our estimator does not rely on a specific response style, and the extracted scenario-based prior remains robust under different prompt formulations.

\paragraph{Additional Evaluation on AlignX.}
To further address the concern regarding dataset sufficiency, we conduct additional experiments on a third dataset, AlignX~\cite{li20251000000users}. 
AlignX provides diverse contextual scenarios, allowing us to evaluate the effectiveness of different methods beyond the original evaluation datasets.
As shown in Tab.~\ref{tab:alignx_results}, BaCVA consistently outperforms all baselines on AlignX, achieving the lowest MAE and the highest correlation and accuracy.
Compared with the strongest baseline, ValuesRAG, BaCVA reduces MAE from 2.184 to 1.872, improves correlation from 0.587 to 0.683, and increases accuracy from 64.40\% to 68.92\%.
These results further confirm the effectiveness of BaCVA on an additional dataset and strengthen the empirical support for our method.

\begin{table}[t]
\centering
\small
\setlength{\tabcolsep}{6pt}
\renewcommand{\arraystretch}{1.08}
\caption{Additional results on AlignX. Best results are in bold.}
\label{tab:alignx_results}
\begin{tabular}{lccc}
\toprule
Method & MAE $\downarrow$ & Corr $\uparrow$ & Acc $\uparrow$ \\
\midrule
Value Prompt & 2.682 & 0.297 & 61.96\% \\
COUPLE & 2.577 & 0.356 & 59.89\% \\
MOD & 2.406 & 0.544 & 64.26\% \\
MetaAligner & 2.442 & 0.370 & 61.29\% \\
PAD & 2.535 & 0.348 & 60.07\% \\
ValuesRAG & 2.184 & 0.587 & 64.40\% \\
\textbf{BaCVA} & \textbf{1.872} & \textbf{0.683} & \textbf{68.92\%} \\
\bottomrule
\end{tabular}
\end{table}

\paragraph{Robustness across Value Systems}
We further evaluate BaCVA on \textsc{PRISM} under an alternative value representation based on the five-dimensional Moral Foundations taxonomy. Compared with the strongest baseline, ValuesRAG, our method consistently achieves performance improvements across evaluation metrics. This consistent gain demonstrates that BaCVA effectively generalizes across different value systems and is capable of robustly modeling personalized alignment signals beyond a specific value taxonomy.

\begin{figure}[t]
    \centering
        \includegraphics[width=\linewidth]{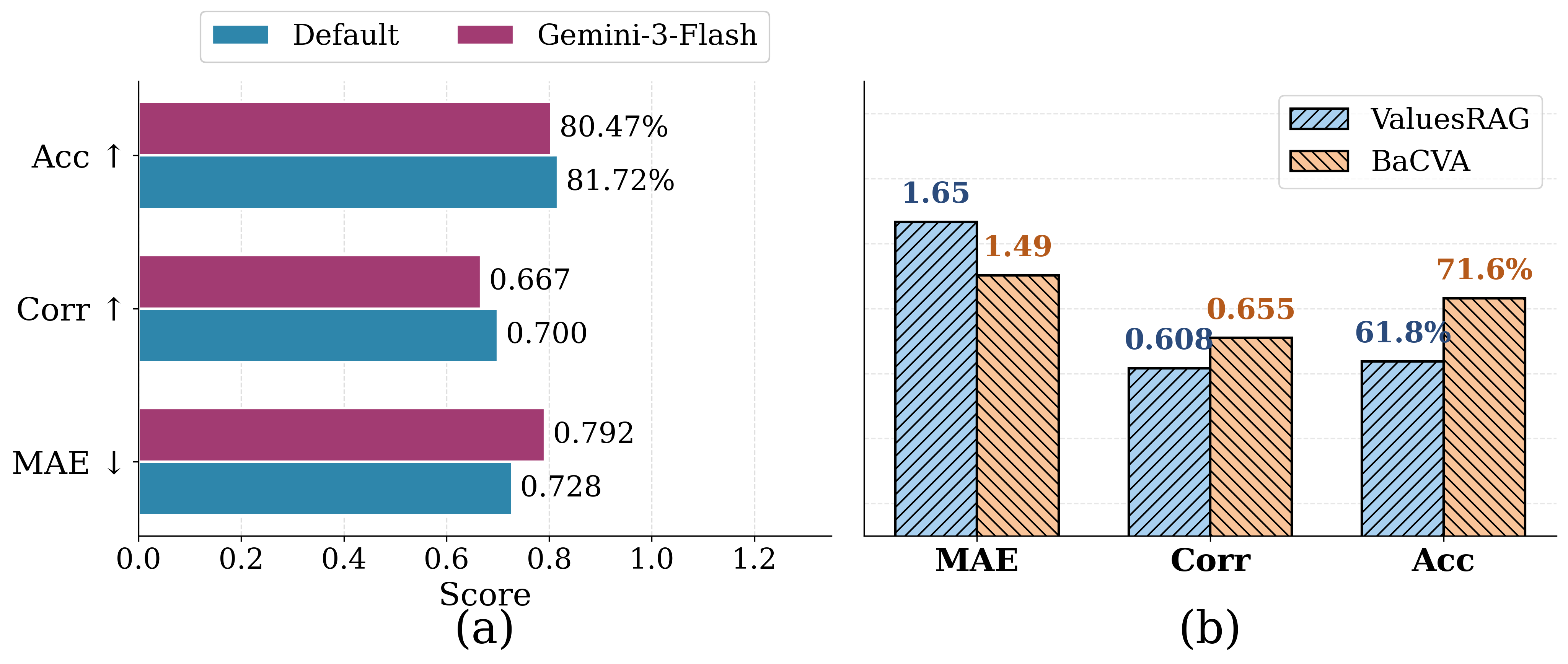}
    \caption{Hyperparameter analysis on universal-response model selection and value-system generalization.}
    \label{fig:hyper}
\end{figure}

\paragraph{Application to Fine-Grained Value Dimensions.}
To further examine whether BaCVA is limited by the granularity of the underlying value taxonomy, we conduct an additional experiment on \textsc{PRISM} using the value system from the Daily-Dilemma dataset~\citep{chiu2024dailydilemmas}, which contains 301 fine-grained value dimensions, such as trust, fairness, and patience. This setting directly tests whether the proposed framework can capture more contextual and nuanced value signals beyond coarse-grained taxonomies.

\begin{table}[t]
\centering
\small
\caption{Results on \textsc{PRISM} under the fine-grained value system from Daily-Dilemma.}
\label{tab:fine_grained_values}
\resizebox{\linewidth}{!}{%
\begin{tabular}{lccc}
\toprule
\textbf{Method} & \textbf{MAE} $\downarrow$ & \textbf{Corr.} $\uparrow$ & \textbf{Accuracy} $\uparrow$ \\
\midrule
PersonValue & 1.325 & 0.109 & 55.74\% \\
ValuesRAG   & 0.717 & 0.167 & 62.63\% \\
BaCVA       & \textbf{0.365} & \textbf{0.471} & \textbf{78.69\%} \\
\bottomrule
\end{tabular}%
}
\end{table}

As shown in Tab.~\ref{tab:fine_grained_values}, BaCVA substantially outperforms both PersonValue and ValuesRAG under the fine-grained value system. These results suggest that BaCVA is not constrained to a fixed coarse-grained taxonomy, but can also adapt to more detailed value dimensions and better capture contextual nuances in personalized value alignment.

\subsection{Cluster-level adaptability analysis}
\label{appendix:cluster}

Tab.~\ref{tab:cluster_analysis} compares BaCVA with two prior-only variants, where Prior-$v_u$ uses only the static user value prior and Prior-$v_x$ uses only the scenario-level value prior. 
Across all five major clusters, BaCVA consistently achieves the lowest MAE, indicating that its posterior value estimation can adapt to diverse contextual topics rather than relying on either user- or scenario-level priors alone.

\begin{table}[t]
\centering
\small
\renewcommand{\arraystretch}{1.2}
\setlength{\tabcolsep}{3pt}
\resizebox{\linewidth}{!}{
\begin{tabular}{l l c c c}
\toprule
\textbf{Cluster / Topic} & \textbf{Main Value Dims.} 
& \textbf{Prior-$v_u$ MAE} $\downarrow$
& \textbf{Prior-$v_x$ MAE} $\downarrow$
& \textbf{BaCVA MAE} $\downarrow$ \\
\midrule
C1 / Social Values & Self-Dir., Benev. 
& 1.539 & 1.986 & \textbf{0.777} \\
C2 / Life Advice & Self-Direction 
& 1.536 & 2.205 & \textbf{0.741} \\
C3 / Politics/Civic & Power, Achievement 
& 2.367 & 2.106 & \textbf{1.158} \\
C4 / Religion/Tradition & Security, Tradition 
& 1.887 & 2.100 & \textbf{0.774} \\
C5 / Relationships & Benevolence, Security 
& 1.380 & 1.890 & \textbf{0.672} \\
\bottomrule
\end{tabular}
}
\caption{Cluster-level adaptability analysis on \textsc{PRISM}.}
\label{tab:cluster_analysis}

\end{table}

\subsection{Case Study}
\begin{figure}[t]
    \centering
    \includegraphics[width=\linewidth]{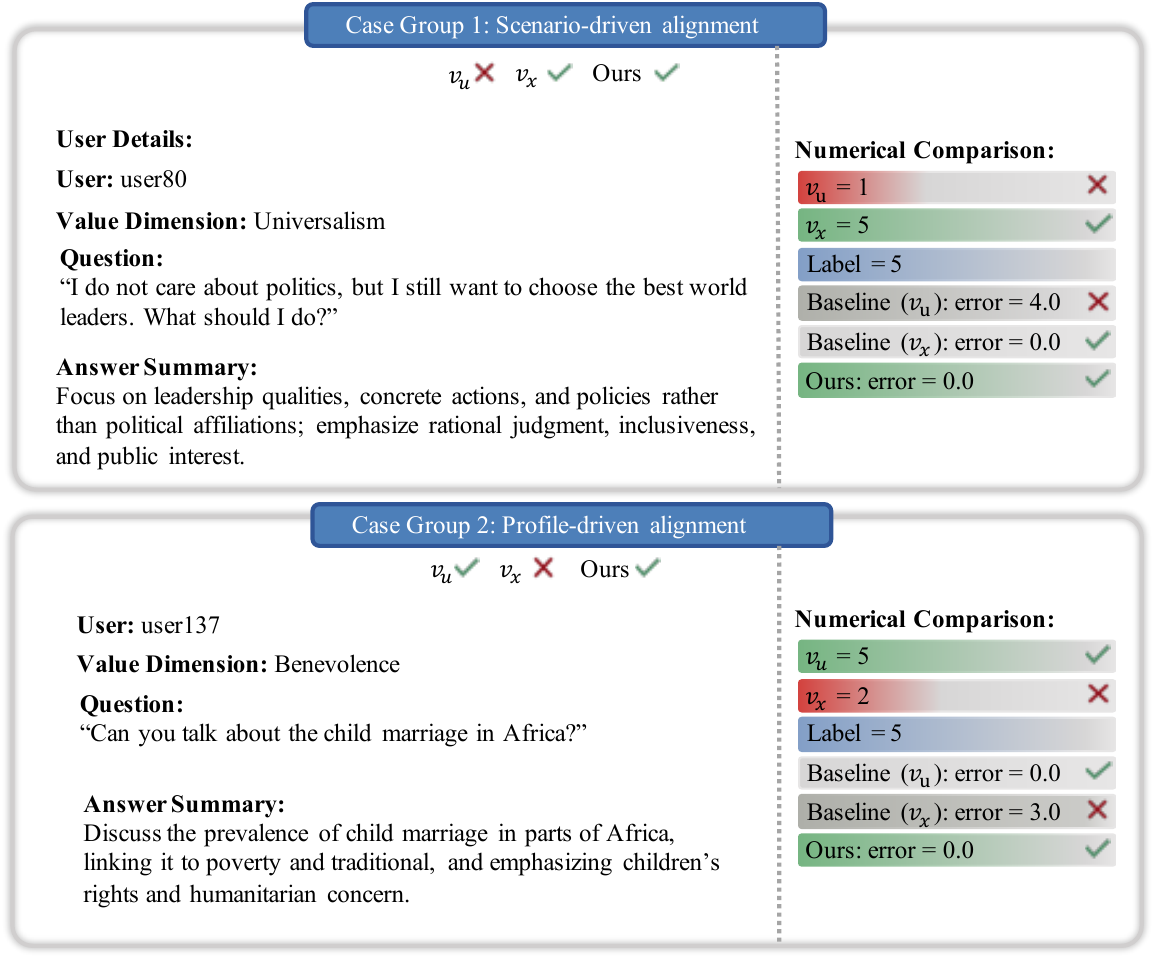}
    \caption{Case study on the PRISM dataset. The proposed model generates responses that are aligned with both user-specific preferences and contextual scenarios, demonstrating its ability to achieve Contextualized Personalization Alignment.}
    \label{fig:case_study}
\end{figure}

We conduct a case study on the PRISM dataset to qualitatively examine whether our model can achieve both \emph{scenario alignment} and \emph{personalization alignment}. As illustrated in Fig.~\ref{fig:case_study}, our model is able to generate responses that are consistent with individual user preferences while simultaneously adapting to different contextual scenarios. This demonstrates that the proposed approach effectively integrates contextual understanding with personalized modeling, achieving \emph{Contextualized Personalization Alignment}.

\subsection{Downstream detail analysis}
\label{appendix:e2e}

\textbf{Framework Formulation.} We extend the two-stage BaCVA framework by coupling contextual value inference with response generation. Following the standard notation, each training instance consists of a user $u$, a question $x$, a user profile $\mathbf{v}_u$, question-level universal values $\mathbf{v}_x$, and question-specific personalized values $\mathbf{v}_{u,x}$. The dataset is augmented with preference triples $(x, y_c, y_r)$, where $y_c$ and $y_r$ denote the human-preferred and dispreferred responses, respectively. In the baseline \textbf{Two-stage BaCVA}, the inferrer predicts $\hat{\mathbf{v}}_{u,x}$ via dual-stream gating, and a decoupled generator subsequently conditions on a discrete text description of $\hat{\mathbf{v}}_{u,x}$. In our proposed \textbf{BaCVA-E2E}, we interface these modules via continuous soft prompts. A linear projector maps the inferred vector $\hat{\mathbf{v}}_{u,x} \in [1,5]^{10}$ to $k$ prefix embeddings $\mathbf{e}_v \in \mathbb{R}^{k \times d}$, which are prepended to the token embeddings of $[x, y]$. The joint training objective is formulated as $\mathcal{L}_{\text{E2E}} = \mathcal{L}_{\text{value}} + \lambda \mathcal{L}_{\text{gen}}$, where $\mathcal{L}_{\text{value}}$ is the masked value-regression loss from the vanilla BaCVA, and $\mathcal{L}_{\text{gen}}$ is the preference alignment loss. This continuous path enables response-quality feedback to backpropagate directly into the inferrer.

\textbf{Preference Alignment Loss.} Conditioned on the soft prompts $\mathbf{e}_v$, we optimize the generator $\pi_\theta$ using a DPO-style objective with multi-line alignment:
\begin{equation}
\begin{split}
\mathcal{L}_{\text{gen}} = -\mathbb{E}_{(x,y_c,y_r)} \Big[ \log \sigma \big( \beta [ &\Delta \log \pi(y_c) - \\ 
& \Delta \log \pi(y_r) ] \big) \Big],
\end{split}
\end{equation}
where the likelihood margin for response $y \in \{y_c, y_r\}$ is defined as:
\begin{equation}
\Delta \log \pi(y) = \log \frac{\pi_\theta(y \mid \mathbf{e}_v, x)}{\pi_{\text{ref}}(y \mid \mathbf{e}_v, x)}.
\end{equation}
and $\beta$ scales the preference margin. Likelihood computations are restricted to answer tokens, masking out prefix and question tokens. We set the reference policy $\log \pi_{\text{ref}} \equiv 0$, $\lambda = 1.0$, and $\beta = 0.1$ across all experiments.

\textbf{Conditioning Protocol.} During training, value vectors are injected as continuous soft prompts ($k=8$) prepended to the LM input, and the generator is fine-tuned via LoRA. Conversely, during inference, to isolate the performance gains attributable to improved value inference rather than a modified decoding interface, BaCVA-E2E adopts the exact same \textbf{discrete text prompting} as the two-stage baseline. The E2E-trained inferrer outputs $\hat{\mathbf{v}}_{u,x}$, which is converted into natural language value levels and inserted into the system prompt (e.g., \emph{Achievement: high (4.2)}). The E2E generator LoRA weights are disabled at test time, and preprocessing follows the stage-1 implementation.

\textbf{Dataset and Preprocessing.} We utilize the \textbf{PRISM} dataset \citep{kirk2024prism} with an 80/20 user-level train/test split (using random seed 42), resulting in a test set of $N=420$. The ground-truth vector $\mathbf{v}_{y_c}$ is derived directly from the 10-dimensional annotation of the human-chosen answer. 

\textbf{Evaluation Metrics.} To comprehensively assess our framework, we evaluate performance across three metrics, each computed exclusively over the answer-specific active dimensions:
\begin{itemize}
    \item \textbf{MAE:} Quantifies the absolute intensity mismatch between the predicted $\hat{v}^d_{u,x}$ and ground-truth $v^d_{y_c}$ across relevant dimensions, formulated as $\text{MAE} = \mathbb{E}_d[|\hat{v}^d_{u,x} - v^d_{y_c}|]$.
    \item \textbf{Spearman's $\rho$:} Measures the rank correlation between $\hat{\mathbf{v}}_{u,x}$ and $\mathbf{v}_{y_c}$ pooled across the entire test set to capture ordinal consistency.
    \item \textbf{Accuracy (Pairwise Ranking):} Evaluates whether the predicted vector is closer to the chosen response than the rejected alternative in terms of $L_1$ distance, defined as $\|\hat{\mathbf{v}}_{u,x} - \mathbf{v}_{y_c}\|_1 < \|\hat{\mathbf{v}}_{u,x} - \mathbf{v}_{y_r}\|_1$. Ties are scored as $0.5$.
\end{itemize}

\textbf{Baseline Configurations.} We compare our proposed framework against two major categories of alignment methods, all sharing the same backbone LM and utilizing greedy decoding ($\text{max\_new\_tokens}=256$). The first category consists of \textbf{non-personalized baselines}, represented by \texttt{DirectAnswer}, which generates responses directly from the user query without any value conditioning. The second category comprises \textbf{personalized baselines}, including:
\begin{itemize}
\item \textbf{ValuePrompt}, which hard-codes the static user profile into the system prompt;

\item \textbf{COUPLE}, which leverages counterfactual prompting to highlight conflicts between $\mathbf{v}_u$ and $\mathbf{v}_x$;

\item \textbf{MetaAligner}, which employs a critique-and-rewrite mechanism to align a draft response with user values;

\item \textbf{PAD} and \textbf{MOD}, which implement contrastive decoding via token-level logit manipulation;

\item \textbf{ValuesRAG}, which augments the generation context with few-shot exemplars retrieved from historically similar profiles.
\end{itemize}

\textbf{Hyperparameters and Ablations.} The complete hyperparameter configuration for training BaCVA-E2E is detailed in Tab.~\ref{tab:e2e_impl}. For ablation analysis, we maintain the identical pipeline but evaluate under two restricted settings: \emph{w/o gradient to inferrer}, where $\hat{\mathbf{v}}_{u,x}$ is detached before soft-prompt projection to isolate decoupled training behavior, and \emph{w/o $\mathcal{L}_{\text{value}}$}, which completely removes the explicit value-regression constraints on the inferrer submodule.

\begin{table}[t]
\centering
\footnotesize
\setlength{\tabcolsep}{4pt}
\begin{tabular}{p{3.0cm} p{4.2cm}}
\toprule
\textbf{Component} & \textbf{Setting} \\
\midrule
Backbone LM 
& Qwen3-1.7B (\texttt{bfloat16}) \\

BaCVA Submodule 
& LoRA ($r{=}16$, $\alpha{=}32$), warm-started \\

Generator Module 
& Attention Projection LoRA ($r{=}16$, $\alpha{=}32$) \\

Soft Prompts 
& $k{=}8$ tokens, Linear Projector + LayerNorm \\

Optimizer 
& AdamW (weight decay $=0.01$) \\

Learning Rate 
& $5{\times}10^{-5}$ (Inferrer); 
  $2{\times}10^{-4}$ (Generator + Projector) \\

Training Layout 
& 3 epochs, batch size 1, gradient accumulation 8 \\

Loss Scalers $(\lambda,\beta)$ 
& $1.0,\ 0.1$ \\

Seed 
& 42 \\
\bottomrule
\end{tabular}
\caption{Hyperparameter settings for BaCVA-E2E training.}
\label{tab:e2e_impl}
\end{table}

Tab.~\ref{tab:ablation_results} presents the performance of BaCVA-E2E and its ablation variants on the PRISM test set. The results show that both gradient feedback to the value inferrer and the value supervision objective contribute positively to the final alignment performance.
 
\begin{table}[htbp]
\centering
\footnotesize
\setlength{\tabcolsep}{5pt}
\begin{tabular}{lccc}
\toprule
\textbf{Method} & \textbf{MAE} $\downarrow$ & \textbf{Acc} $\uparrow$ & \textbf{Spearman} $\uparrow$ \\
\midrule
\multicolumn{4}{l}{\textit{Proposed Framework}} \\
BaCVA (Two-stage)         & 0.3835 & 0.7975 & 0.4782 \\
\textbf{BaCVA-E2E}        & \textbf{0.3776} & \textbf{0.7996} & \textbf{0.4812} \\
\midrule
\multicolumn{4}{l}{\textit{Ablations}} \\
\quad w/o grad.\ to inferrer & 0.3853 & 0.7957 & 0.4794 \\
\quad w/o $\mathcal{L}_{\text{value}}$   & 0.3841 & 0.7258 & 0.4757 \\
\bottomrule
\end{tabular}
\caption{Ablation results on the PRISM test set ($N=420$).}
\label{tab:ablation_results}
\end{table}

\begin{figure}[ht]
    \centering
    \includegraphics[width=\columnwidth]{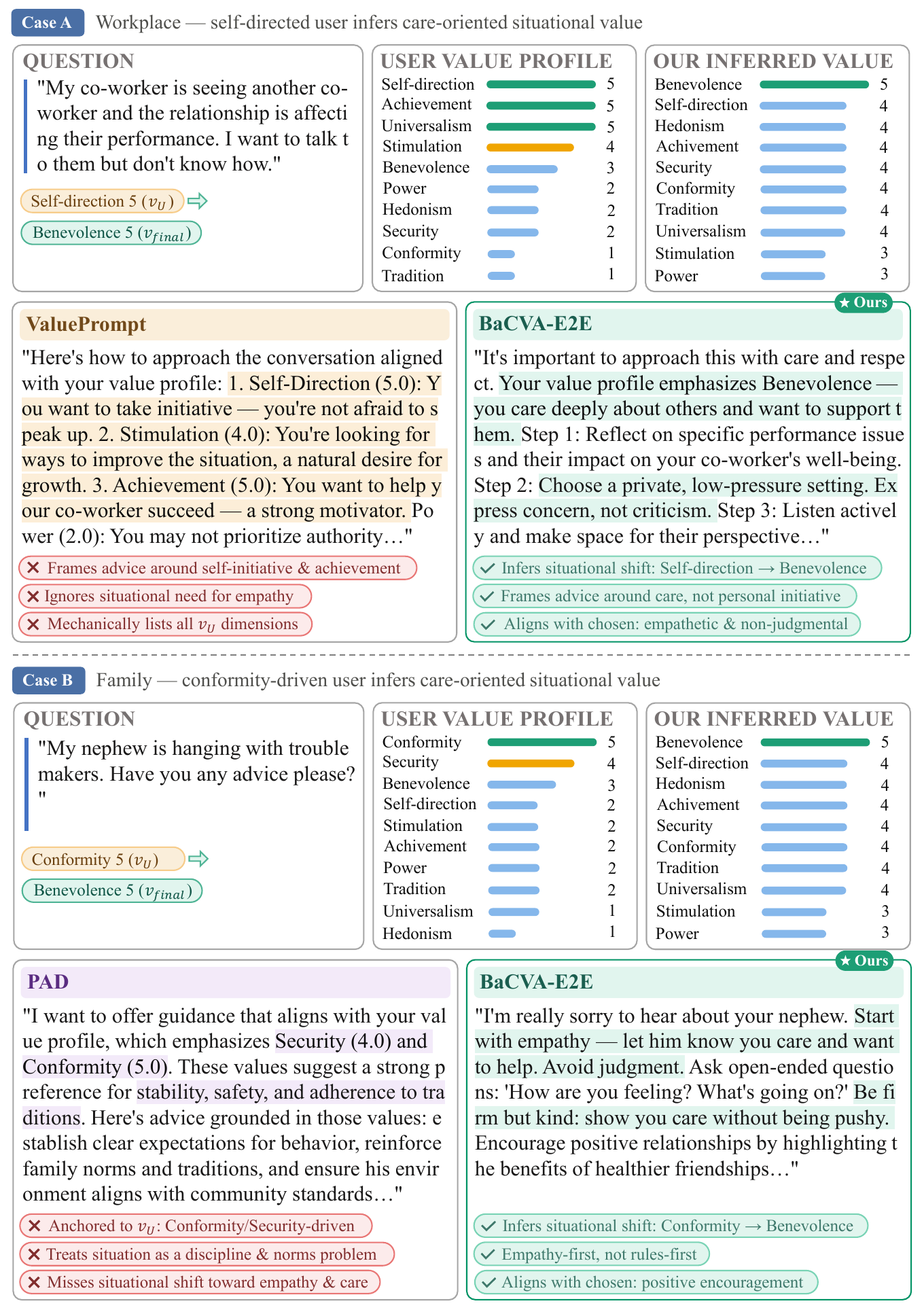}
    \caption{Qualitative comparison of BaCVA-E2E against representative baselines on two PRISM test instances where the situational value $\hat{\mathbf{v}}_{u,x}$ diverges from the static user profile $\mathbf{v}_u$.}
    \label{fig:appendix_cases}
\end{figure}

\textbf{Qualitative Case Studies.}
Fig.~\ref{fig:appendix_cases} presents two cases where the situational value $\hat{\mathbf{v}}_{u,x}$ diverges from the static profile $\mathbf{v}_u$. In Case~A, \textbf{ValuePrompt} frames the response around the user's self-direction-dominant $\mathbf{v}_u$, missing the empathetic tone the situation requires; BaCVA-E2E correctly infers a Benevolence-dominant $\hat{\mathbf{v}}_{u,x}$ and responds accordingly. In Case~B, \textbf{PAD} anchors advice to the user's Conformity-dominant $\mathbf{v}_u$ and treats the situation as a discipline problem; BaCVA-E2E infers that the family context elevates Benevolence and produces empathy-first guidance instead.

\subsection{Cross-Extractor Evaluation}
\label{app:cross_extractor}

The main experiments use GPT-5-nano to construct both training and test value targets.
Although its agreement with human annotations is validated in Appendix~\ref{app:human_eval}, using the same extractor throughout the pipeline may introduce extractor-specific evaluation bias.
We therefore evaluate whether BaCVA's improvements transfer to independently constructed value targets from different model families.

\paragraph{Human Validation of Value Extractors}
We first compare GPT-5-nano, Gemini-3-Flash, and Qwen3-235B-A22B against the same human-annotated value assessments.
As shown in Tab.~\ref{tab:extractor_human_validation}, all three extractors achieve strong agreement with human annotations at both the profile and answer levels.

\begin{table}[t]
\centering
\small
\resizebox{\linewidth}{!}{%
\begin{tabular}{lcccc}
\toprule
\textbf{Extractor}
& \textbf{Profile MAE} $\downarrow$
& \textbf{Profile Corr.} $\uparrow$
& \textbf{Answer MAE} $\downarrow$
& \textbf{Answer Corr.} $\uparrow$ \\
\midrule
GPT-5-nano       & \textbf{0.056} & \textbf{0.986} & \textbf{0.120} & \textbf{0.936} \\
Gemini-3-Flash   & 0.172 & 0.930 & 0.166 & 0.880 \\
Qwen3-235B-A22B  & 0.164 & 0.932 & 0.171 & 0.878 \\
\bottomrule
\end{tabular}%
}
\caption{Human validation of value extractors.}
\label{tab:extractor_human_validation}
\end{table}

\begin{table*}[htbp]
\centering
\small
\setlength{\tabcolsep}{5pt}
\renewcommand{\arraystretch}{1.1}
\begin{tabular}{lllccc}
\toprule
\textbf{Training Extractor}
& \textbf{Test Extractor}
& \textbf{Method}
& \textbf{MAE} $\downarrow$
& \textbf{Corr.} $\uparrow$
& \textbf{Accuracy} $\uparrow$ \\
\midrule
GPT-5-nano & GPT-5-nano
& ValuesRAG & 0.952 & 0.638 & 71.68\% \\
GPT-5-nano & GPT-5-nano
& BaCVA & \textbf{0.728} & \textbf{0.700} & \textbf{81.72\%} \\
\midrule
GPT-5-nano & Gemini-3-Flash
& ValuesRAG & 1.052 & 0.605 & 69.02\% \\
GPT-5-nano & Gemini-3-Flash
& BaCVA & \textbf{0.826} & \textbf{0.665} & \textbf{78.55\%} \\
\midrule
GPT-5-nano & Qwen3-235B-A22B
& ValuesRAG & 1.066 & 0.598 & 68.37\% \\
GPT-5-nano & Qwen3-235B-A22B
& BaCVA & \textbf{0.842} & \textbf{0.657} & \textbf{77.96\%} \\
\bottomrule
\end{tabular}
\caption{Cross-extractor evaluation with mismatched training and test value targets.}
\label{tab:cross_extractor_results}
\end{table*}

\paragraph{Evaluation with Mismatched Training and Test Extractors}
We retain GPT-5-nano targets for training, but independently reconstruct the test targets using Gemini-3-Flash and Qwen3-235B-A22B.
Neither test extractor is used during training or scenario-prior construction.
For each extractor, the values of both preferred and dispreferred test responses are reconstructed before computing MAE, correlation, and pairwise ranking accuracy.

As shown in Tab.~\ref{tab:cross_extractor_results}, BaCVA consistently outperforms ValuesRAG under both matched and mismatched evaluation.
Across the two mismatched settings, BaCVA reduces MAE by 21.2\% on average and improves accuracy by 9.56 percentage points over ValuesRAG, while also achieving higher correlation.

\subsection{Robustness across Scenario-Prior Estimators}
\label{app:prior_estimator_robustness}

To evaluate whether BaCVA depends on a particular model for constructing the scenario prior $\mathbf{v}_x$, we compare three aligned LLMs from different model families: GPT-5-nano, Gemini-3-Flash, and Qwen3-235B-A22B.
These models are used as operational proxies for population-level scenario salience rather than definitive representations of universal human values.

We measure the consistency of the extracted scenario priors using within-one agreement.
For each question and value dimension, two estimators are considered consistent when their scores differ by at most one point on the 1--5 scale.
The agreement reported for each estimator is its average pairwise agreement with the other two estimators over the same PRISM questions.

As shown in Tab.~\ref{tab:prior_estimator_robustness}, the three estimators achieve an average within-one agreement of 89.15\%.
The downstream PRISM MAE has a standard deviation of only 0.027, indicating that BaCVA maintains stable absolute-error performance across different scenario-prior estimators, although some variation remains in correlation and pairwise accuracy.

\begin{table*}[t]
\centering
\small
\setlength{\tabcolsep}{5pt}
\renewcommand{\arraystretch}{1.1}
\begin{tabular}{lcccc}
\toprule
\textbf{Scenario-Prior Estimator}
& \textbf{Prior Agreement} $\uparrow$
& \textbf{MAE} $\downarrow$
& \textbf{Corr.} $\uparrow$
& \textbf{Accuracy} $\uparrow$ \\
\midrule
GPT-5-nano
& 90.20\% & 0.728 & 0.700 & 81.72\% \\
Gemini-3-Flash
& 89.89\% & 0.792 & 0.667 & 80.47\% \\
Qwen3-235B-A22B
& 87.35\% & 0.749 & 0.746 & 79.81\% \\
\midrule
Mean $\pm$ Std.
& 89.15\% $\pm$ 1.28\%
& 0.756 $\pm$ 0.027
& 0.704 $\pm$ 0.032
& 80.67\% $\pm$ 0.79\% \\
\bottomrule
\end{tabular}
\caption{Robustness to the model used for constructing the scenario prior on PRISM. Prior agreement denotes average pairwise within-one agreement on the 1--5 value scale.}
\label{tab:prior_estimator_robustness}
\end{table*}

\subsection{Generalization across Alignment Backbones}
\label{app:backbone_robustness}

To evaluate whether the improvements of BaCVA generalize beyond the default Qwen3-8B backbone, we additionally implement BaCVA and representative personalized baselines using Llama3-8B.
All methods use the same datasets, data splits, value extractor, input information, and baseline configurations.

As shown in Tab.~\ref{tab:backbone_robustness}, BaCVA consistently outperforms ValuePrompt and ValuesRAG under both backbone models.
With Llama3-8B, BaCVA reduces MAE by 37.9\%, improves correlation by 0.125, and increases accuracy by 19.06 percentage points relative to ValuesRAG.
These results support the generalizability of BaCVA across alignment backbone families.

\begin{table}[t]
\centering
\small
\setlength{\tabcolsep}{4pt}
\renewcommand{\arraystretch}{1.1}
\resizebox{\linewidth}{!}{%
\begin{tabular}{llccc}
\toprule
\textbf{Backbone}
& \textbf{Method}
& \textbf{MAE} $\downarrow$
& \textbf{Corr.} $\uparrow$
& \textbf{Accuracy} $\uparrow$ \\
\midrule
\multirow{3}{*}{Qwen3-8B}
& ValuePrompt & 1.790 & 0.568 & 72.95\% \\
& ValuesRAG   & 0.952 & 0.638 & 71.68\% \\
& BaCVA       & \textbf{0.728} & \textbf{0.700} & \textbf{81.72\%} \\
\midrule
\multirow{3}{*}{Llama3-8B}
& ValuePrompt & 2.969 & 0.312 & 42.29\% \\
& ValuesRAG   & 1.524 & 0.369 & 63.02\% \\
& BaCVA       & \textbf{0.947} & \textbf{0.494} & \textbf{82.08\%} \\
\bottomrule
\end{tabular}%
}
\caption{Personalized value alignment results across backbone models on PRISM.}
\label{tab:backbone_robustness}
\end{table}

\subsection{Profile--Answer Conflict Analysis}
\label{app:conflict_analysis}

To examine when contextual value inference is most beneficial, we measure the conflict between the global user profile $\mathbf{v}_u$ and the values expressed in the preferred answer $\mathbf{v}_{y_c}$.
For each sample, the profile--answer distance is computed as
\begin{equation}
d_{\mathrm{conflict}}
=
\frac{1}{D}
\sum_{j=1}^{D}
\left|
v_u^{(j)}-v_{y_c}^{(j)}
\right|,
\end{equation}
where $D$ is the number of value dimensions.
We divide the PRISM test samples into low-, medium-, and high-conflict subsets according to this distance.

As shown in Tab.~\ref{tab:conflict_analysis}, the performance gap between BaCVA and profile-based methods becomes more pronounced as profile--answer conflict increases.
On the high-conflict subset, BaCVA reduces MAE by 17.8\% relative to ValuesRAG.
Its absolute improvement over ValuesRAG is also larger in the medium- and high-conflict subsets than in the low-conflict subset, indicating that explicit contextual value inference is particularly useful when answer-level values diverge from the global user profile.

\begin{table}[t]
\centering
\small
\setlength{\tabcolsep}{4pt}
\renewcommand{\arraystretch}{1.1}
\resizebox{\linewidth}{!}{%
\begin{tabular}{lrrrr}
\toprule
\textbf{Conflict Level}
& \textbf{\#Samples}
& \textbf{ValuePrompt}
& \textbf{ValuesRAG}
& \textbf{BaCVA} \\
\midrule
Low    & 136 & 1.471 & 0.801 & \textbf{0.640} \\
Medium & 149 & 1.705 & 0.899 & \textbf{0.678} \\
High   & 135 & 2.059 & 1.163 & \textbf{0.956} \\
\bottomrule
\end{tabular}%
}
\caption{MAE on PRISM under different degrees of profile--answer conflict.}
\label{tab:conflict_analysis}
\end{table}

\subsection{General-Purpose Capabilities}
\label{app:general_capability}

To assess whether personalized value-alignment training degrades capabilities outside the target task, we evaluate the original Qwen3-8B backbone and BaCVA on four general-purpose benchmarks: BoolQ, ARC-Easy, HellaSwag, and WinoGrande~\citep{clark2019boolq,clark2018think,zellers2019hellaswag,sakaguchi2021winogrande}.
These benchmarks cover reading comprehension, science reasoning, commonsense completion, and commonsense reasoning.

As shown in Tab.~\ref{tab:general_capability}, BaCVA maintains performance comparable to the original backbone across all four benchmarks.
Its average performance is 78.74\%, compared with 77.69\% for the original backbone, indicating that BaCVA training does not substantially degrade general-purpose capabilities.

\begin{table}[hbtp]
\centering
\small
\setlength{\tabcolsep}{5pt}
\renewcommand{\arraystretch}{1.1}
\begin{tabular}{lcc}
\toprule
\textbf{Benchmark}
& \textbf{Backbone}
& \textbf{BaCVA} \\
\midrule
BoolQ      & 86.70\% & 86.45\% \\
ARC-Easy   & 81.06\% & 83.59\% \\
HellaSwag  & 74.96\% & 76.87\% \\
WinoGrande & 68.03\% & 68.03\% \\
\midrule
Average    & 77.69\% & 78.74\% \\
\bottomrule
\end{tabular}
\caption{General-purpose performance of the original Qwen3-8B backbone and BaCVA after personalized value-alignment training.}
\label{tab:general_capability}
\end{table}

\subsection{Comparison with Explicit Stochastic Approximation}
\label{app:posterior_approximation}

To examine the accuracy--efficiency trade-off of the practical posterior implementation, we compare BaCVA with an explicit stochastic approximation that draws 10 Monte Carlo samples for each input.
The two implementations use the same backbone, test set, and evaluation protocol.

As shown in Tab.~\ref{tab:posterior_approximation}, the amortized implementation reduces MAE by 7.4\%, improves correlation by 0.021, and increases accuracy by 3.97 percentage points relative to the 10-sample Monte Carlo approximation.
Its inference latency is 7.3\% higher but remains comparable under the same evaluation setting.
These results support the use of deterministic amortized inference as a practical approximation rather than exact stochastic posterior computation.

\begin{table}[htbp]
\centering
\small
\setlength{\tabcolsep}{4pt}
\renewcommand{\arraystretch}{1.1}
\resizebox{\linewidth}{!}{%
\begin{tabular}{lcccc}
\toprule
\textbf{Implementation}
& \textbf{MAE} $\downarrow$
& \textbf{Corr.} $\uparrow$
& \textbf{Accuracy} $\uparrow$
& \textbf{Time} $\downarrow$ \\
\midrule
10-sample Monte Carlo
& 0.786 & 0.679 & 77.75\% & \textbf{148.0 ms} \\
Amortized BaCVA
& \textbf{0.728} & \textbf{0.700} & \textbf{81.72\%} & 158.8 ms \\
\bottomrule
\end{tabular}%
}
\caption{Accuracy--efficiency comparison of posterior implementations on PRISM. Inference time is reported per sample.}
\label{tab:posterior_approximation}
\end{table}

\section{Generative Assistance}

Generative AI tools were used to assist with language polishing and improving the clarity of presentation in this paper. All technical content, experimental design, results, and conclusions were developed and verified by the authors.

\end{document}